\documentclass[11pt]{article}

\usepackage[final]{acl}

\usepackage{times}
\usepackage{latexsym}
\usepackage{caption}
\usepackage{mathtools}
\usepackage{amsmath,amssymb}

\usepackage{subcaption}
\usepackage[T1]{fontenc}
\usepackage[table]{xcolor}
\definecolor{lightblue}{rgb}{0.93, 0.97, 1.0}

\usepackage[utf8]{inputenc}

\usepackage{microtype}

\usepackage{inconsolata}

\usepackage{float}

\usepackage{multirow}
\usepackage{graphicx}
\usepackage{booktabs}
\usepackage[most]{tcolorbox}   
\usepackage{xcolor}
\newcommand{\hlg}[1]{%
  \colorbox{green!12}{\parbox{\dimexpr\linewidth-2\fboxsep\relax}{\raggedright #1}}%
}
\newcommand{\hlr}[1]{%
  \colorbox{red!10}{\parbox{\dimexpr\linewidth-2\fboxsep\relax}{\raggedright #1}}%
}
\usepackage{longtable,booktabs}
\usepackage{array}
\newcommand{\method}{\textsc{ThinkBrake}}

\makeatletter
\newcommand{\customfootnote}[1]{%
  \begingroup
  \renewcommand\thefootnote{}%
  \footnotetext{#1}%
  \addtocounter{footnote}{0}%
  \endgroup
}
\makeatother

\title{ThinkBrake: Efficient Reasoning via Log-Probability Margin \\ Guided Decoding}

\author{
  \textbf{Sangjun Song\textsuperscript{$*$}},
  \textbf{Minjae Oh\textsuperscript{$*$}},
  \textbf{Seungkyu Lee\textsuperscript{$\ddagger$}},
  \textbf{Sungmin Jo\textsuperscript{$\S$}},
  \textbf{Yohan Jo\textsuperscript{$\dagger$}}
\\
  Graduate School of Data Science, Seoul National University
  \\
    \texttt{\{ssangjun706, kosair, gyuu01, jsm0424, yohan.jo\}@snu.ac.kr}
}

\begin{document}
\maketitle
\begin{NoHyper}
\customfootnote{
\\[-1.0em]
\noindent \textsuperscript{$*$}Equal contribution \quad 
\textsuperscript{$\dagger$}Corresponding author. \\
\textsuperscript{$\ddagger$}Visiting intern from Department of Industrial Engineering, Seoul National University. \\
\textsuperscript{$\S$}Visiting intern from Department of Computer Science and Engineering, Seoul National University.}
\end{NoHyper}


\begin{abstract}
Large Reasoning Models (LRMs) allocate substantial inference-time compute to Chain-of-Thought (CoT) reasoning, improving performance on mathematics, scientific QA, and tool usage. However, this introduces \emph{overthinking}: LRMs often reach a correct intermediate solution, continue reasoning, and overwrite it with an incorrect answer. We first demonstrate that oracle stopping—where we inject \texttt{</think>} at every sentence boundary and select the best stopping point in hindsight—improves average accuracy by 8\% while reducing thinking tokens by 72\%, exposing substantial overthinking. Motivated by this finding, we propose \method{}, which monitors the log-probability margin between the top continuation token and \texttt{</think>} at sentence boundaries, stopping reasoning when this margin narrows. \method{} requires no training and achieves favorable accuracy–efficiency trade-offs across math, scientific QA, and tool usage benchmarks, reducing thinking token usage by up to 30\%. Furthermore, we provide theoretical analysis showing that \method{} is equivalent to test-time realignment with a reward bonus for the \texttt{</think>}. Code is available at \url{https://github.com/holi-lab/ThinkBrake}.
\end{abstract}


\section{Introduction}
\label{sec:intro}
\begin{figure*}[t]
    \centering
    \includegraphics[width=\textwidth]{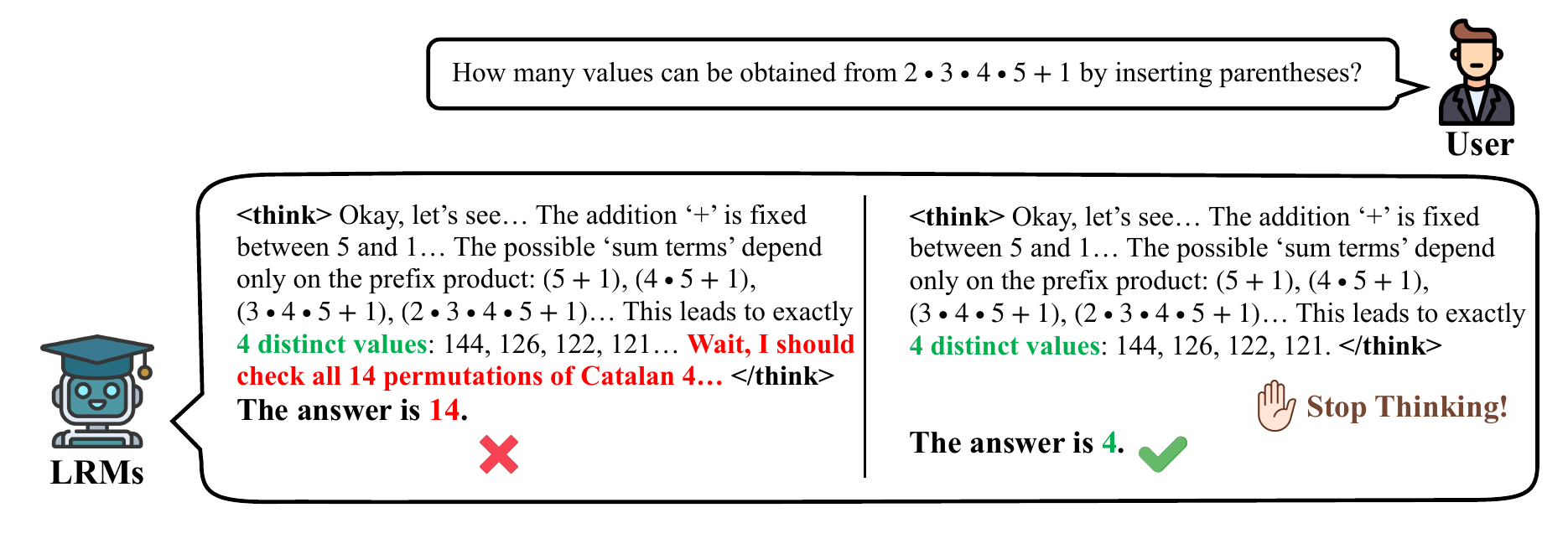}
    \caption{An example of LRM overthinking (left). The LRM arrives at the correct solution of 4, but continues reasoning after ``Wait'' and overwrites it with an incorrect answer of 14. Appropriate early termination via \texttt{</think>} injection removes overthinking and leads to the correct answer (right).}
    \label{fig:overthinking_example}
\end{figure*}

Recent progress in Large Reasoning Models (LRMs) \citep{OpenAI,guo2025deepseek, yang2025qwen3} has demonstrated remarkable capabilities across various tasks such as mathematics, scientific QA, and tool usage. By allocating inference-time computation through Chain-of-Thought (CoT) reasoning \citep{wei2022chain}, LRMs demonstrate behaviors like self-correction and iterative refinement, actively improving performance with increased token usage \citep{chen2025towards, xu2025survey}. However, this increased inference-time compute raises practical concerns about \emph{efficient reasoning} \citep{feng2025efficient}, as longer trajectories introduce higher latency and costs. More critically, it introduces \textit{overthinking} \citep{zhang2025making}: LRMs frequently reach a correct intermediate step, only to continue deliberating and overwrite it with an incorrect final output (see Figure~\ref{fig:overthinking_example}, left panel).

We first conduct a preliminary oracle analysis to validate that overthinking is a genuine problem and that LRMs would benefit from early termination of reasoning. Leveraging the fact that recent LRMs use special tokens \texttt{<think>} and \texttt{</think>} to wrap reasoning (where \texttt{</think>} triggers answer generation), we inject \texttt{</think>} at every sentence boundary, forcing early termination at each possible reasoning step (see \S\ref{sec:oracle}). Across mathematical reasoning benchmarks (GSM8K, MATH500, AIME2024, AIME2025) and tool usage (Berkeley Function Calling Leaderboard; BFCL), this oracle approach achieves an average accuracy gain of 8\% while reducing reasoning tokens by 72\%. This gap reveals substantial recoverable headroom: many failures can be prevented simply by stopping at the right moment (see Figure~\ref{fig:overthinking_example}, right panel).

Motivated by this oracle analysis, we introduce \method{}, a practical approach to early stopping. \method{} monitors the log-probability margin between \texttt{</think>} and the current top token at sentence boundaries, stopping reasoning when this margin narrows. \method{} is model-agnostic, requiring only an explicit reasoning format with \texttt{</think>}. While prior works have explored test-time methods for reasoning efficiency \citep{li2025thinkless,wang2025wait,laaouach2025haltcot}, they often rely on hand-crafted heuristics. In contrast, we provide a theoretical foundation for \method{}, showing it is equivalent to KL-regularized test-time realignment that assigns a reward bonus to \texttt{</think>}, pushing the LRM towards concise reasoning (see \S\ref{sec:theory}).

We evaluate \method{} on six LRMs across four mathematical reasoning benchmarks (GSM8K, MATH500, AIME2024, AIME2025), two additional reasoning benchmarks (GPQA-Diamond, ARC-Challenge), and two tool usage benchmarks (BFCL, Meta-Tool), demonstrating consistent efficiency gains with competitive accuracy (see \S\ref{sec:experiments}). We show that \method{} preserves accuracy uniformly across problem difficulties, rather than trading off performance on easy versus hard instances. Furthermore, \method{}-generated trajectories can serve as training data for Direct Preference Optimization (DPO). This enables LRMs to internalize concise reasoning without inference-time intervention and achieve improved efficiency with only a small amount of training data (see \S\ref{sec:training}).  Our contributions are:  

\begin{itemize}
    \item We identify and quantify overthinking in LRMs via oracle \texttt{</think>} rollouts, quantifying recoverable headroom across mathematical reasoning and tool usage (see \S\ref{sec:oracle}).
    \item We introduce \method{}, a test-time inference method based on a log-margin criterion at sentence boundaries, with theoretical grounding as KL-regularized test-time realignment assigning a reward bonus to \texttt{</think>} (see \S\ref{sec:method}).
    \item We validate \method{} across six LRMs and eight benchmarks spanning mathematical, scientific, and tool-use reasoning, and show its generated data enables efficient reasoning via DPO (see \S\ref{sec:experiments}).
\end{itemize}


\section{Overthinking in LRMs}
\label{sec:oracle}
\begin{figure*}[t]
    \centering
    \includegraphics[width=\textwidth]{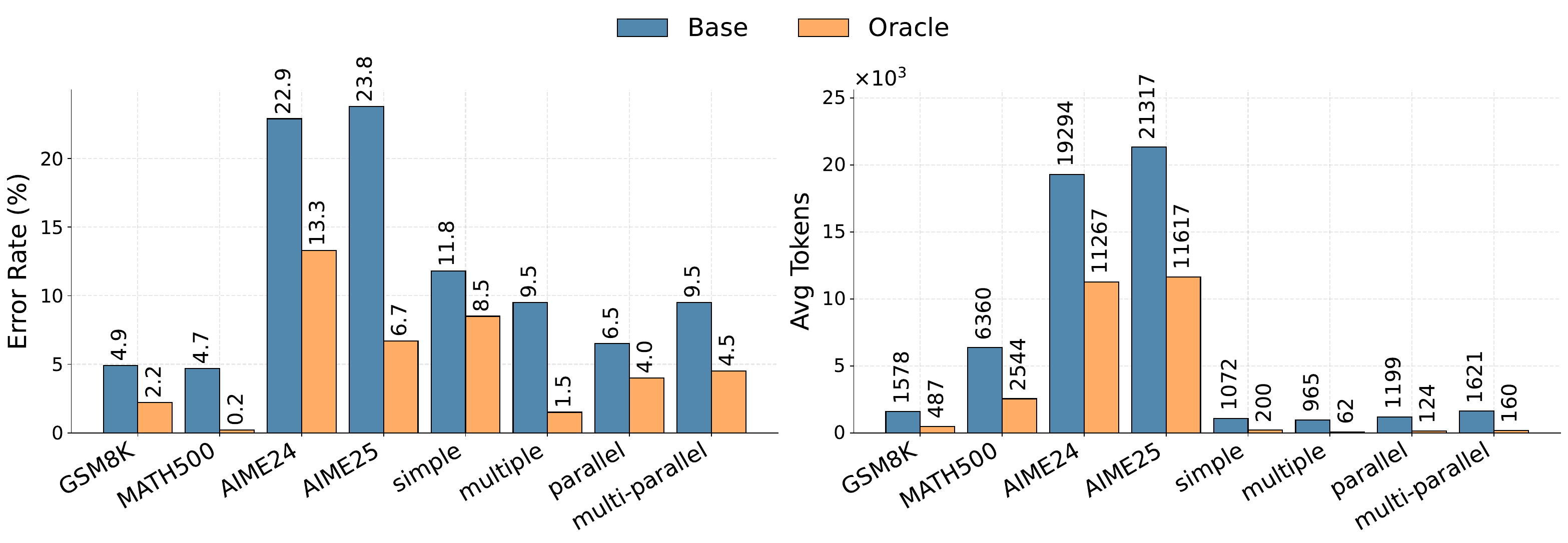}
    \caption{Oracle experiments via sentence-boundary \texttt{</think>} injection on mathematical reasoning (GSM8K, MATH500, AIME2024, AIME2025) and tool usage (BFCL-v1). Optimal termination yields both improved error rate (left) and reduced thinking token usage (right). See Table~\ref{tab:early_stop_split} for a tabular version.}
    \label{fig:oracle_experiment}
\end{figure*}

To validate that overthinking is a genuine problem and quantify the potential gains from early stopping, we analyze whether LRMs would have answered correctly had reasoning been terminated earlier. We conduct controlled rollouts with forced termination by injecting \texttt{</think>} at every sentence boundary, causing the model to stop thinking and produce an answer. We mark a trajectory as \emph{recoverable} if it contains a sentence boundary where the model has reached the correct answer but continues reasoning and eventually produces an incorrect one. This yields an \emph{oracle} accuracy—the achievable accuracy with hindsight knowledge of the optimal stopping point (see Appendix~\ref{app:overthinking_example} for an example).

We use Qwen3-4B-Thinking \citep{yang2025qwen3} as our base LRM and evaluate across mathematical reasoning (GSM8K \citep{cobbe2021gsm8k}, MATH500 \citep{hendrycks2021measuring}, AIME2024, AIME2025 \citep{AoPS_AIME}) and tool usage (BFCL-v1 \citep{patil2025the}). Detailed experiment settings are in Appendix~\ref{app:exp_settings}.

In Figure~\ref{fig:oracle_experiment}, oracle stopping reduces error rate to 6\% on mathematical reasoning and 5\% on tool usage, improving over the baseline by 8\% while reducing thinking token usage by 54\% on mathematics and 89\% on tool usage. Remarkably, the improvements are consistent across difficulty levels; even on the challenging AIME benchmarks, oracle stopping recovers 61\% of failures while cutting tokens by 44\%. Notably, fewer than 5\% of errors are irrecoverable for most benchmarks, confirming that a substantial portion of failures stem from overthinking rather than inability. 
\section{\method{}}
\label{sec:method}
\subsection{Intuition and Definition}
\label{sec:method_subsec}
We propose \method{}, a simple early termination method that injects the \texttt{</think>} when the log-probability margin between the top-predicted token and \texttt{</think>} becomes small. Formally, let $\pi_\theta$ be our LRM, $x$ be the input, and $y_{<t}$ be the current reasoning chain. At each sentence boundary, let $y^\star_t=\arg\max_{y} \pi_\theta(y\mid x;y_{<t})$ be the top-predicted token that is not \texttt{</think>}. We terminate model thinking when:
\begin{align}
    \label{eq:TB}
    \log\frac{\pi_\theta(y_t^{\star}\mid x; y_{<t})} {\pi_\theta(\texttt{</think>}\mid x;y_{<t})} \leq \tau,
\end{align}
where $\tau$ is a hyperparameter that controls the stopping threshold. The intuition is that as productive reasoning concludes, the probability margin between continuing and stopping narrows. While \method{} considers only the top token, it naturally accounts for different levels of overall token entropy, as we justify through the following exhaustive cases:

\paragraph{Case 1: Termination state.} The model assigns the highest probability to \texttt{</think>} ($p_\theta(\texttt{</think>})>p_\theta(y^\star_t)$), so termination occurs regardless of \method{}.

\paragraph{Case 2: High-entropy state.} The model spreads probability across many tokens, indicating genuine uncertainty. Early stopping here risks premature termination of productive reasoning, so we should stop as conservatively as possible. Using $y^\star_t$ as the reference is the most conservative choice, as it by definition carries the largest probability.

\paragraph{Case 3: Low-entropy state.} The model concentrates probability on one or a few tokens, indicating high raw probability of $y^\star_t$. If \texttt{</think>} is comparably probable, it too carries high probability--meaning the model genuinely considers stopping as an option and productive reasoning has ended. Termination in this case is therefore natural.

For comparison, we also consider a linear probability-gap variant (\method{}-p) that triggers when $p_\theta(y^\star_t) - p_\theta(\texttt{</think>}) \leq \tau_\text{prob}$. This variant proves significantly less effective (see \S\ref{sec:results_logprob}). Furthermore, we validate the dynamics of \method{} with respect to entropy and compare its difference against \method{}-p in Appendix~\ref{app:entropy}.

\subsection{Theoretical Analysis}
\label{sec:theory}
We theoretically ground \method{} by showing it is equivalent to a logit margin test, which can be interpreted as test-time realignment via a KL-regularized policy with a reward bonus for \texttt{</think>}.
\paragraph{Log-space margins equal logit margins.} We first show that \method{} is equivalent to a \emph{logit margin} test. For brevity, let $e \coloneqq \texttt{</think>}$ denote the stop token and $s_t \coloneqq (x; y_{<t})$ denote the decoding state. Let $z_\theta(y \mid s_t)$ be the pre-softmax logit for token $y$ in the vocabulary $\mathcal{V}$ at state $s_t$. With decoding temperature $T>0$, the next-token distribution is:
\begin{align}
    \pi_\theta(y \mid s_t)
    \;=\;
    \frac{\exp\!\big(z_\theta(y \mid s_t)/T\big)}{\sum_{y'} \exp\!\big(z_\theta(y' \mid s_t)/T\big)}.
\end{align}
Taking the log-ratio between any two tokens cancels the normalization:
\begin{align}
    \log\frac{\pi_\theta(a \mid s_t)}{\pi_\theta(b \mid s_t)}
    \;=\;
    \frac{z_\theta(a \mid s_t)-z_\theta(b \mid s_t)}{T}.
    \label{eq:logit_margin}
\end{align}
Let $y_t^\star \;\coloneqq\; \arg\max_{y \in \mathcal{V}\setminus\{e\}} z_\theta(y \mid s_t)$ be the best continuation token (excluding $e$). The \emph{log-space stopping margin} is
\begin{align}
    m_t \coloneqq \log\frac{\pi_\theta(y_t^\star \mid s_t)}{\pi_\theta(e \mid s_t)} 
    = \frac{z_\theta(y_t^\star \mid s_t)-z_\theta(e \mid s_t)}{T}.
    \label{eq:thinkbrake_margin}
\end{align}
Therefore, \method{} is exactly a temperature-scaled \emph{logit margin test} between the best continuation token and \texttt{</think>}.

\paragraph{\method{} as test-time realignment.}
We now formalize \method{} as optimizing a KL-regularized policy that prefers emitting \texttt{</think>}. Let $\mathcal{B}$ denote the set of sentence boundary states where \method{} may terminate. At a boundary state $s_t\in\mathcal{B}$, consider the general KL-regularized objective with hyperparameter $\beta > 0$:

\begin{align}
    \tilde{\pi}_{\theta}(\cdot\mid s_t)
    = \arg&\max_{\pi(\cdot\mid s_t)}
    \mathbb{E}_{y\sim \pi(\cdot\mid s_t)}[r(s_t,y)]
    \nonumber\\
    &- \beta\,\mathrm{KL}(\pi(\cdot\mid s_t) \| \pi_\theta(\cdot\mid s_t)).
    \label{eq:kl_obj}
\end{align}

The closed-form optimal policy is given by \citep{korbak2022on, rafailov2023direct}:
\begin{align}
\tilde{\pi}_{\theta}(y\mid s_t)
=
\frac{1}{Z_t}\,
\pi_\theta(y\mid s_t)\,
\exp\!\left(\frac{r(s_t,y)}{\beta}\right).
\label{eq:exp_tilt}
\end{align}

We define the reward to make the KL-regularized objective equivalent to \method{}'s logit margin test in Eq.~\eqref{eq:TB}.
\begin{align}
r_\tau(s_t,y)
\;\coloneqq\;
\tau \beta \cdot \mathbf{1}[y=e].
\label{eq:reward_bonus}
\end{align}
Plugging Eq.~\eqref{eq:reward_bonus} into Eq.~\eqref{eq:exp_tilt} gives $\exp(r_\tau/\beta)=\exp(\tau\,\mathbf{1}[y=e])$, so the realignment simply upweights \texttt{</think>} by a factor $e^\tau$.

We now show that setting $\beta = T$ yields a direct connection to logit margins. Since $\pi_\theta(\cdot\mid s_t)$ is a softmax over logits, Eq.~\eqref{eq:exp_tilt} with $\beta = T$ is equivalent to \emph{logit shaping}. Define the shaped logit:
\begin{align}
\tilde z_{\theta,\tau}(y\mid s_t) &\coloneqq z_\theta(y\mid s_t) + r_\tau(s_t,y), \label{eq:shaped_logit}
\end{align}
where $r_\tau(s_t, y) = \tau T \cdot \mathbf{1}[y=e]$ under our choice of $\beta = T$. Then the realigned policy becomes:
\begin{align}
\tilde{\pi}_{\theta,\tau}(y\mid s_t) &= \mathrm{softmax}(\tilde z_{\theta,\tau}(y\mid s_t)/T).
\label{eq:shaped_policy}
\end{align}
Thus, at boundary states, the realigned policy is exactly the base model with a constant bonus $\tau T$ added to the \texttt{</think>} logit.

\paragraph{Equivalence to the \method{} margin test.}
Under greedy decoding from the realigned policy at $s_t \in \mathcal{B}$, it selects $e$ if and only if
\begin{align}
z_\theta(e\mid s_t)+\tau T &\ge z_\theta(y_t^\star\mid s_t) \nonumber\\
\tau &\geq \frac{z_\theta(y_t^\star\mid s_t)-z_\theta(e\mid s_t)}{T} \nonumber\\
m_t &\le \tau,
\label{eq:threshold_equiv}
\end{align}
where $m_t$ is defined in Eq.~\eqref{eq:thinkbrake_margin}. This is exactly \method{}.
For non-boundary states $s_t\notin\mathcal{B}$, \method{} applies no realignment and decodes from $\pi_\theta(\cdot\mid s_t)$ as usual.

\begin{table*}[t]
\centering
\small
\setlength{\tabcolsep}{3pt}
\renewcommand{\arraystretch}{1.03}
\begin{tabular}{lcccccccccccccc}
\toprule
 & \multicolumn{2}{c}{GSM8K} & \multicolumn{2}{c}{MATH500} & \multicolumn{2}{c}{AIME2024} & \multicolumn{2}{c}{AIME2025} & \multicolumn{2}{c}{GPQA-D} & \multicolumn{2}{c}{ARC-C} & \multicolumn{2}{c}{Avg.}\\
\cmidrule(lr){2-3}\cmidrule(lr){4-5}\cmidrule(lr){6-7}\cmidrule(lr){8-9}\cmidrule(lr){10-11}\cmidrule(lr){12-13}\cmidrule(lr){14-15}
Model
& Acc & $\Delta$Tok & Acc & $\Delta$Tok & Acc & $\Delta$Tok & Acc & $\Delta$Tok & Acc & $\Delta$Tok & Acc & $\Delta$Tok & Acc & $\Delta$Tok \\
\midrule
Qwen3-4B-Thinking & 95.1 & -- & 95.3 & -- & 77.1 & -- & 76.2 & -- & 64.1 & -- & 94.3 & -- & 83.7 & -- \\
\quad + NoWait  & 94.2 & -25.1\% & 96.4 & -17.9\% & 66.7 & -30.5\% & 66.7 & -35.6\% & 68.2 & -19.6\% & 92.7 & 14.6\% & 80.8 & -19.0\%\\
\quad + ThinkLess  & 92.6 & -100\% & 56.4 & -100\% & 3.3 & -100\% & 3.3 & -100\% & 50.0 & -100\% & 94.2 & -100\% & 50.0 & -100\%\\
\quad + DEER & 95.1 & -34.1\% & 94.4 & -14.2\% & 63.3 & -34.3\% & 60.0 & -37.9\% & 68.2 & -1.3\% & 92.8 & -9.0\% & 79.0 & -21.8\% \\
\quad + Dynasor-CoT & 95.2 & -34.0\% & 87.2 & -42.1\% & 63.3 & -28.8\% & 53.3 & -28.9\% & 62.6 & -22.6\% & 94.5 & -39.1\% & 76.0 & -23.0\% \\
\quad + \method{}-p  & 93.9 & -37.3\% & 91.2 & -57.9\% & 35.0 & -72.6\% & 30.4 & -76.0\% & 47.0 & -67.5\% & 93.7 & -31.7\% & 65.2 & -57.2\%\\
\rowcolor{lightblue}
\quad + \method{} & 94.8 & -18.9\% & 96.6 & -20.4\% & 67.2 & -33.5\% & 62.8 & -33.4\% & 62.6 & -32.7\% & 94.0 & -19.9\% & 79.7 & -26.5\% \\
\midrule
Qwen3-4B  & 94.4 & -- & 96.0 & -- & 68.8 & -- & 59.6 & -- & 51.0 & -- & 94.0 & -- & 77.3 & --\\
\quad + NoWait  & 94.5 & -40.4\% & 94.6 & -31.4\% & 56.7 & -16.6\% & 50.0 & -27.1\% & 57.6 & -26.6\% & 93.4 & -14.4\% & 74.5 & -26.1\%\\
\rowcolor{lightblue}
\quad + \method{}  & 94.5 & -30.1\% & 95.4 & -14.4\% & 64.6 & -12.7\% & 60.4
& -13.5\% & 55.6 & -29.7\% & 93.9 & -11.5\% & 77.4 & -18.7\%\\
\midrule
Qwen3-14B  & 96.0 & -- & 96.8 & -- & 71.7 & -- & 69.2 & -- & 60.6 & -- & 95.9 & -- & 81.7 & --\\
\quad + NoWait  & 95.8 & -29.4\% & 95.2 & -25.1\% & 70.0 & -19.8\% & 56.7 & -16.6\% & 62.1 & -20.5\% & 96.2 & -8.1\% & 79.3 & -19.9\%\\
\rowcolor{lightblue}
\quad + \method{}  & 95.0 & -15.4\% & 96.9 & -7.7\% & 77.9 & -7.4\% & 65.8 & -6.0\% & 61.1 & -39.9\% & 95.2 & -18.2\% & 82.0 & -15.8\%\\
\midrule
Qwen3-32B  & 96.0 & -- & 97.0 & -- & 75.8 & -- & 68.3 & -- & 65.2 & -- & 92.1 & -- & 82.4 & --\\
\quad + NoWait  & 95.3 & -23.6\% & 95.8 & -20.6\% & 66.7 & -12.0\% & 56.7 & -10.6\% & 60.1 & -12.4\% & 90.1 & 9.0\% & 77.5 & -11.7\%\\
\rowcolor{lightblue}
\quad + \method{}  & 96.5 & -9.1\% & 97.2 & -1.4\% & 77.1 & -6.2\% & 68.3 & -4.3\% & 65.7 & -19.9\% & 91.0 & -8.0\% & 82.6 & -8.2\%\\
\midrule
DeepSeek-R1-7B & 92.7 & -- & 93.8 & -- & 49.5 & -- & 37.6 & -- & 48.0 & -- & 67.7 & -- & 64.9 & -- \\
\quad + NoWait  & 90.6 & -30.8\% & 91.2 & -31.0\% & 40.0 & -32.4\% & 26.7 & -36.5\% & 43.4 & -35.5\% & 64.6 & -12.6\% & 59.4 & -29.8\%\\
\rowcolor{lightblue}
\quad + \method{} & 92.4 & -25.0\% & 92.0 & -23.2\% & 45.2 & -25.0\% & 33.2 & -20.3\% & 45.5 & -51.2\% & 70.7 & -35.3\% & 63.2 & -30.0\% \\
\midrule
Phi-4-Reasoning  & 91.7 & -- & 71.1 & -- & 67.5 & -- & 63.7 & -- & 62.6 & -- & 80.0 & -- & 72.8 & --\\
\quad + NoWait  & 92.1 & -2.4\% & 70.8 & -17.7\% & 63.3 & -9.4\% & 53.3 & -8.6\% & 63.1 & -19.1\% & 77.6 & -7.0\% & 70.0 & -10.7\%\\
\rowcolor{lightblue}
\quad + \method{}  & 91.7 & -25.6\% & 70.4 & -17.2\% & 73.3 & -6.4\% & 53.3 & -5.2\% & 67.7 & -1.1\% & 80.0 & -25.3\% & 72.7 & -13.5\%\\
\bottomrule
\end{tabular}
\caption{Math and science results on GSM8K, MATH500, AIME2024, AIME2025, GPQA-D, and ARC-C. We report accuracy and $\Delta$Tok (thinking token reduction vs. the base decoding) for each model. For AIME2024 and AIME2025, which contain only 30 questions, we report results over 32 runs and also report the variance and confidence intervals. For brevity we only show the baseline method for the Qwen3-4B-Thinking model. A comprehensive result table including baseline methods is in Appendix~\ref{app:full}.}

\label{tab:math_results}
    \vspace{-12pt}
\end{table*}

\section{Experiments}
\label{sec:experiments}
\subsection{Experimental Setup}
\label{sec:exp_setup}
We evaluate several models for our main experiments: Qwen3-4B-Thinking, Qwen3-4B, Qwen3-14B, Qwen3-32B \citep{yang2025qwen3}, DeepSeek-R1-7B \citep{guo2025deepseek}, and Phi-4-Reasoning \citep{abdin2025phi}. Qwen3-32B is a model post-trained with reinforcement learning (RL), while the smaller Qwen models are trained via supervised fine-tuning. DeepSeek-R1-7B is a Qwen2.5-based model distilled from DeepSeek-R1-generated reasoning traces. Phi-4-Reasoning is derived from Phi-4 through supervised fine-tuning on curated reasoning demonstrations. All models use default reasoning format with explicit \texttt{<think>} and \texttt{</think>} delimiters.

We compare against NoWait \citep{wang2025wait}, which suppresses filler tokens like "wait" and "hmm"; ThinkLess \citep{li2025thinkless}, which forces immediate \texttt{</think>} emission after \texttt{<think>}; Dynasor-CoT \citep{fu2025reasoning}, which extracts intermediate answers and applies early stopping based on consistency; DEER \citep{yang2025dynamic}, which also interrupts mid-thinking and stops based on entropy; and \method{}-p (\S\ref{sec:method_subsec}). Details in Appendix~\ref{app:baseline}.

We report both task accuracy and thinking token reduction. We measure \emph{thinking tokens} as tokens generated within the reasoning span (between \texttt{<think>} and \texttt{</think>}). We report $\Delta$Tok as the percentage change in thinking tokens relative to baseline decoding for the same model and benchmark. Detailed baseline methods, implementation settings, and system prompts are provided in Appendix~\ref{app:exp_settings}. 

\label{sec:selecting_hyperparam}
\begin{figure}[!t]
    \centering
    \includegraphics[width=\columnwidth]{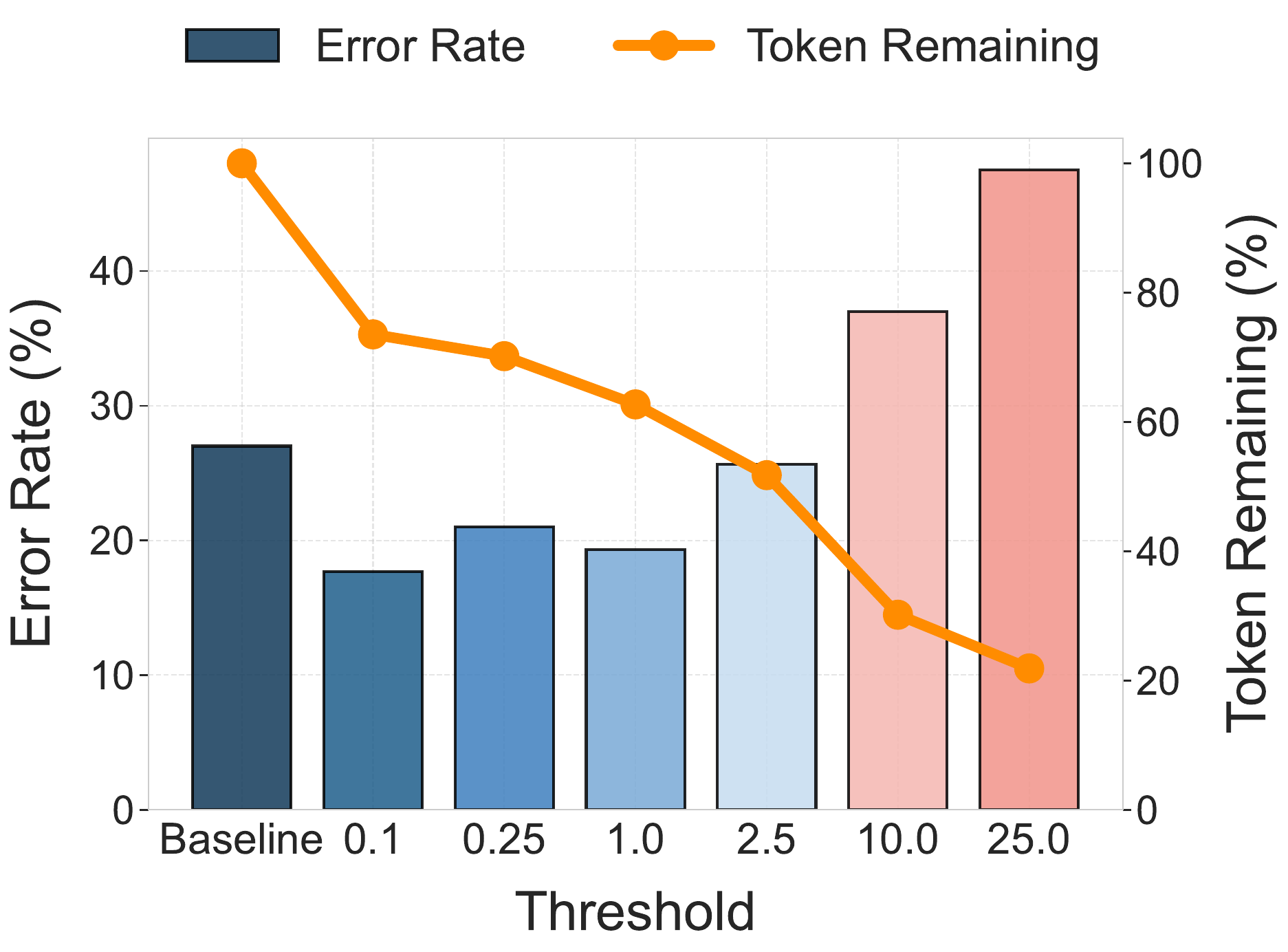}
    \caption{$\tau$ search from error rate and thinking token usage for \method{} on Qwen3-4B-Thinking.}
    \label{fig:tau_select}
    \vspace{-12pt}
\end{figure}
\paragraph{Selecting Hyperparameter $\tau$.}
\method{} introduces a single hyperparameter, $\tau$, which controls how aggressively we terminate reasoning at sentence boundaries. Increasing $\tau$ encourages earlier emission of \texttt{</think>}, yielding fewer thinking tokens, but overly large values can truncate useful reasoning and degrade accuracy. To choose a default threshold without per-benchmark tuning, we perform a lightweight sweep over $\tau$ on a small validation split. Concretely, we evaluate Qwen3-4B-Thinking across a grid of $\tau$ values on a subset of the DAPO17K \citep{yu2025dapo} dataset and record both error rate and thinking token reduction compared to baseline decoding. 

Figure~\ref{fig:tau_select} shows that at $\tau{=}0.1$, \method{} achieves both reduced error rate and substantial token reduction compared to baseline. Beyond this point, error rate increases and token usage decreases smoothly as $\tau$ increases further. Notably, while error rate is lowest at $\tau{=}0.1$, it remains competitive for $\tau$ values up to 1.0, demonstrating robustness to the choice of threshold. We therefore select $\tau{=}0.1$ and apply it to all Qwen3 model sizes and all benchmarks. Our results show that this threshold generalizes well out-of-distribution from mathematical reasoning to other tasks. Similarly, we select optimal $\tau$ values for DeepSeek-R1-7B ($\tau{=}0.1$) and Phi-4-Reasoning ($\tau{=}2.5$) using the same procedure (see Appendix~\ref{app:hyperparameter_selection} for details).

\begin{table*}[t]
\centering
\small
\setlength{\tabcolsep}{3pt}
\renewcommand{\arraystretch}{1.03}
\resizebox{\textwidth}{!}{
\begin{tabular}{lcccccccccccccc}
\toprule
& \multicolumn{4}{c}{BFCL-v1} & \multicolumn{4}{c}{BFCL-v2} & \multicolumn{4}{c}{Meta-Tool} & \multicolumn{2}{c}{} \\
\cmidrule(lr){2-5}\cmidrule(lr){6-9}\cmidrule(lr){10-13}
Model & \multicolumn{2}{c}{Parallel} & \multicolumn{2}{c}{Multi-Parallel} & \multicolumn{2}{c}{Parallel} & \multicolumn{2}{c}{Multi-Parallel} & \multicolumn{2}{c}{Single} & \multicolumn{2}{c}{Multiple} & \multicolumn{2}{c}{Avg.} \\
\cmidrule(lr){2-3}\cmidrule(lr){4-5}\cmidrule(lr){6-7}\cmidrule(lr){8-9}\cmidrule(lr){10-11}\cmidrule(lr){12-13}\cmidrule(lr){14-15}
& Acc & $\Delta$Tok & Acc & $\Delta$Tok & Acc & $\Delta$Tok & Acc & $\Delta$Tok & Acc & $\Delta$Tok & Acc & $\Delta$Tok & Acc & $\Delta$Tok \\
\midrule
Qwen3-4B-Thinking & 93.5 & -- & 90.5 & -- & 87.5 & -- & 79.2 & -- & 69.7 & -- & 85.5 & -- & 84.3 & -- \\
\quad + NoWait & 88.5 & -44.4\% & 88.5 & -53.5\% & 81.3 & -50.7\% & 75.0 & -66.5\% & 68.9 & 0.1\% & 84.7 & -11.2\% & 81.2 & -37.7\% \\
\quad + ThinkLess & 79.5 & -100\% & 83.0 & -100\% & 37.5 & -100\% & 45.8 & -100\% & 67.4 & -100\% & 86.5 & -100\% & 66.6 & -100\% \\
\quad + \method{}-p & 78.0 & -26.9\% & 42.5 & -33.4\% & 62.5 & -28.5\% & 29.2 & -43.8\% & 70.8 & -25.8\% & 86.7 & -12.7\% & 61.6 & -28.5\% \\
\rowcolor{lightblue}
\quad + \method{} & 95.5 & -26.8\% & 90.5 & -28.2\% & 87.5 & -32.3\% & 87.5 & -18.9\% & 72.4 & -20.6\% & 86.3 & -9.0\% & 86.6 & -22.6\% \\
\midrule
Qwen3-4B & 90.0 & -- & 83.0 & -- & 81.2 & -- & 70.8 & -- & 74.3 & -- & 90.5 & -- & 81.6 & -- \\
\quad + NoWait & 88.0 & -13.5\% & 84.0 & 1.2\% & 75.0 & -5.1\% & 75.0 & -11.1\% & 72.5 & -3.1\% & 90.5 & -8.4\% & 80.8 & -6.7\% \\
\rowcolor{lightblue}
\quad + \method{} & 89.5 & -16.5\% & 83.5 & -12.0\% & 62.5 & -21.1\% & 75.0 & -10.2\% & 72.7 & -1.4\% & 91.1 & -3.0\% & 79.0 & -10.7\% \\
\midrule
Qwen3-14B & 92.0 & -- & 83.5 & -- & 62.5 & -- & 70.8 & -- & 63.3 & -- & 84.9 & -- & 76.2 & -- \\
\quad + NoWait & 90.0 & -16.3\% & 84.0 & 1.8\% & 62.5 & -30.1\% & 62.5 & -10.9\% & 61.2 & -2.4\% & 84.5 & -5.8\% & 74.1 & -10.6\% \\
\rowcolor{lightblue}
\quad + \method{} & 93.0 & -21.2\% & 85.0 & -14.2\% & 62.5 & -25.4\% & 62.5 & -7.6\% & 66.3 & -1.6\% & 85.7 & -4.5\% & 75.8 & -12.4\% \\
\midrule
Qwen3-32B & 93.0 & -- & 85.0 & -- & 68.8 & -- & 70.8 & -- & 64.8 & -- & 84.3 & -- & 77.8 & -- \\
\quad + NoWait & 91.0 & -13.0\% & 82.5 & -5.0\% & 68.8 & -3.9\% & 62.5 & -14.3\% & 63.8 & 3.6\% & 83.9 & 1.3\% & 75.4 & -5.2\% \\
\rowcolor{lightblue}
\quad + \method{} & 91.0 & -14.3\% & 90.0 & -7.3\% & 68.8 & -4.6\% & 58.0 & -13.3\% & 64.4 & -1.5\% & 85.5 & -1.1\% & 76.3 & -7.0\% \\
\midrule
DeepSeek-R1-7B & 52.5 & -- & 38.0 & -- & 43.8 & -- & 29.2 & -- & 63.2 & -- & 80.3 & -- & 51.2 & -- \\
\quad + NoWait & 62.0 & -4.0\% & 42.5 & -6.9\% & 50.0 & -8.9\% & 20.8 & 11.0\% & 64.9 & 0.6\% & 73.2 & -19.3\% & 52.2 & -4.6\% \\
\rowcolor{lightblue}
\quad + \method{} & 42.0 & -14.6\% & 37.0 & -11.6\% & 43.8 & -19.9\% & 16.7 & -6.0\% & 63.4 & -7.4\% & 76.3 & -26.3\% & 46.5 & -14.3\% \\
\midrule
Phi-4-Reasoning & 87.5 & -- & 77.5 & -- & 81.2 & -- & 58.3 & -- & 77.7 & -- & 90.5 & -- & 78.8 & -- \\
\quad + NoWait & 81.0 & -1.3\% & 75.5 & -13.2\% & 81.2 & -6.4\% & 58.3 & -11.9\% & 76.6 & -8.5\% & 90.9 & -15.4\% & 77.2 & -9.5\% \\
\rowcolor{lightblue}
\quad + \method{} & 84.5 & 2.7\% & 77.5 & -6.4\% & 75.0 & 4.6\% & 70.8 & -4.9\% & 77.2 & -2.5\% & 90.3 & -5.4\% & 79.2 & -2.0\% \\
\bottomrule
\end{tabular}}
\caption{Results on BFCL-v1, BFCL-v2, and Meta-Tool benchmarks. Simple and Multiple scenarios in BFCL are omitted for brevity. We report accuracy and $\Delta$Tok (token reduction) for parallel and multi-parallel scenarios in BFCL, and single and multiple selection scenarios in Meta-Tool. A comprehensive result table including all baseline methods is in Appendix~\ref{app:full}.}
\label{tab:tool_results}
\vspace{-10pt}
\end{table*}

\subsection{Main Results}
\label{sec:main_results}
\paragraph{Math \& Science Reasoning.}
We evaluate on four math benchmarks: GSM8K \citep{cobbe2021gsm8k}, MATH500 \citep{hendrycks2021measuring}, AIME2024, AIME2025 \citep{AoPS_AIME}. We also consider a scientific reasoning benchmark and a general reasoning benchmark: GPQA-D \citep{rein2024gpqa}, ARC-Challenge \citep{clark2018thinksolvedquestionanswering}. For AIME benchmarks, we provide 32 run results with confidence intervals for Qwen3-4B-Thinking and DeepSeek-R1-7B (see Appendix~\ref{app:more_results}).

Table~\ref{tab:math_results} shows that \method{} consistently achieves strong accuracy-efficiency trade-offs across all models and difficulties. On easier benchmarks (GSM8K, MATH500), \method{} reduces tokens by 10--30\% while maintaining or improving accuracy. On challenging AIME benchmarks, \method{} maintains competitive performance with substantial token savings—notably improving accuracy for Qwen3-14B and Phi-4-Reasoning on AIME2024, suggesting \method{} prevents overthinking. 

In contrast, baseline methods show various limitations. ThinkLess (which removes all thinking) causes catastrophic drops on hard math: $\sim$40\% drop on MATH500 and $\sim$73\% drop on AIME2024 for Qwen3-4B-Thinking, highlighting the need for reasoning in such tasks. NoWait degrades less severely but still drops notably on challenging AIME compared to \method{}. Dynasor-CoT shows high performance degradation on benchmarks like AIME; while it often reduces more tokens, this comes at the cost of substantially larger performance degradation. DEER shows competitive performance but still degrades on benchmarks like AIME compared to \method{}.


An apparent pattern across model scales is that larger models require less aggressive stopping. Average token reduction decreases from 26.5\% (Qwen3-4B-Thinking) to 15.8\% (14B) to 8.2\% (32B). To verify this trend of diminishing gains as model size scales, we further evaluate on a larger model: Qwen3-Next-80B. Surprisingly, we observe a substantial 22.3\% token reduction, breaking the apparent trend. We therefore attribute the diminishing reductions from 4B to 32B not to model size, but to reasoning trace length, as larger models tend to produce shorter traces, leaving less redundant thinking to prune. The 80B model, despite being the largest, generates longer traces than the 32B model and correspondingly yields greater reductions. See Appendix~\ref{app:80B} for a further analysis.

\paragraph{Tool Usage.}
\label{sec:results_tool}
We evaluate on BFCL non-live (v1), live (v2) \citep{patil2025the} and Meta-Tool \citep{huang2024metatool}. Tool usage results show different patterns, as ThinkLess—which failed catastrophically on math—performs reasonably on some simpler tool usage scenarios, suggesting function calling benefits less from extended reasoning than mathematical domains.

However, on challenging scenarios, baseline heuristics can degrade performance. On BFCL-v2 with Qwen3-4B-Thinking, ThinkLess suffers catastrophic drops on parallel tasks ($\sim$50\% drop) and multi-parallel tasks ($\sim$33\% drop). NoWait shows modest degradation with a $\sim$7\% drop on parallel and $\sim$5\% drop on multi-parallel. In contrast, \method{} maintains robust performance across all scenarios. On BFCL-v2, \method{} matches baseline accuracy on parallel tasks with 32.3\% token reduction, and notably improves multi-parallel performance by $\sim$8\%. This demonstrates that \method{}'s adaptive stopping criterion successfully prevents overthinking on complex tool usage scenarios where simpler heuristics fail. These results show that \method{} generalizes effectively beyond mathematical and scientific reasoning to tool usage tasks.

\paragraph{Logarithmic Margin Design.}
\label{sec:results_logprob}
As discussed in \S\ref{sec:method_subsec}, we compare \method{} with its linear probability-gap variant \method{}-p: $p_\theta(y^\star_t) - p_\theta(\texttt{</think>}) \leq \tau_{\text{prob}}$. Tables~\ref{tab:math_results} and~\ref{tab:tool_results} show that \method{}-p exhibits substantial accuracy drops on challenging tasks like AIME--achieving only 30--35\% on AIME compared to \method{}'s 62--67\%—while performing comparably on easier tasks like GSM8K with Qwen3-4B-Thinking.

This is likely because the log-ratio formulation requires the absolute probabilities of both $y^\star_t$ and \texttt{</think>} to be sufficiently large to trigger termination. When both probabilities are low, a small log-margin is harder to achieve—the two must be far more similar in relative terms than when both are high. As a result, \method{} terminates model thinking conservatively when the model is uncertain, and primarily when it is confident (see Appendix~\ref{app:log_explanation}). In contrast, raw probability differences lack this adaptive property, causing premature termination during uncertain reasoning states and leading to significant accuracy degradation on challenging tasks. These results validate that the log-space formulation is essential for \method{}'s robustness on hard reasoning tasks.

\paragraph{Premature Exiting and Spurious Reasoning.}
While \method{} largely preserves baseline accuracy, we observe slight performance degradation on highly complex benchmarks (e.g., AIME, GPQA-D) for specific models such as Qwen3-4B-Thinking and DeepSeek-R1-7B. To determine whether this stems from premature termination, a vulnerability in confidence-based early-exit methods, we compare \method{}'s stopping points against an oracle setting. As shown in Table~\ref{tab:token_length_results}, \method{} consistently generates more tokens than the oracle before exiting, indicating that it does not suffer from premature termination in practice, and is using sufficient reasoning budgets. 

Consequently, we attribute these accuracy drops not to insufficient reasoning, but to \emph{spurious reasoning}. Because harder benchmarks naturally elicit longer reasoning traces, they pose a greater risk for the model to engage in flawed or contradictory logic \textit{after} successfully solving the problem, but only before \method{} triggers termination. This explains why the observed degradation is primarily concentrated in benchmarks characterized by extended reasoning lengths. We provide further results and a qualitative example of spurious reasoning in Appendix~\ref{app:spurious}.

\begin{figure*}[!t]
    \centering
    \includegraphics[width=\textwidth]{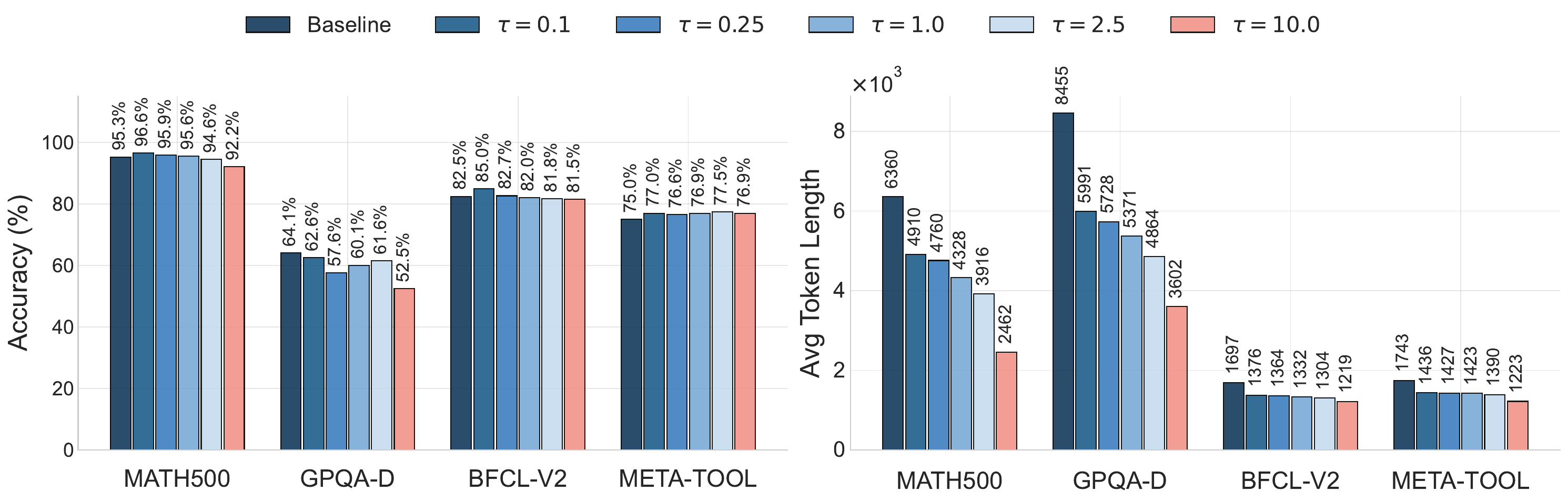}
    \caption{Hyperparameter sensitivity for Qwen3-4B-Thinking over MATH500, GPQA-D and tool usages. Accuracy remains robust except for extreme case $\tau = 10$ (left). Token reduction increases with $\tau$ (right).}
    \label{fig:hyperparameter_experiment}
\end{figure*}

\begin{table}[t]
\centering
\small
\begin{tabular}{lcc}
\toprule
Benchmark & Oracle & \method{} \\
\midrule
AIME2024 & 11,261 & 12,830 \\
AIME2025 & 11,618 & 14,197 \\
\bottomrule
\end{tabular}
\caption{Average tokens for Oracle and \method{} on Qwen3-4B-Thinking. \method{} exits later than the oracle for both AIME2024 and AIME2025.}
\label{tab:token_length_results}
\vspace{-15pt}
\end{table}

\subsection{Ablation Studies}
\label{sec:ablation}

\paragraph{Hyperparameter Sensitivity.}
We further analyze \method{} across test sets with various thresholds ($\tau \in \{0.1, 0.25, 1.0, 2.5, 10\}$) for Qwen3-4B-Thinking. Figure~\ref{fig:hyperparameter_experiment} shows that accuracy remains stable until very high thresholds ($\tau \geq 10$), while token reduction increases monotonically with $\tau$. Although a single threshold works reasonably well across various tasks within model families, optimal thresholds vary by task difficulty. Easier tasks like MATH500 and tool usages maintain accuracy with higher $\tau$, while challenging GPQA-D shows sensitivity to increasing $\tau$, suggesting we can use more aggressive $\tau$ for easier tasks to maximize token reduction. As a practical rule of thumb, $\tau = 0.1$ works well across almost all model-benchmark pairs; for new settings, one can increase $\tau$ for easier tasks to maximize token efficiency or maintain a low $\tau$ for harder ones to preserve accuracy. See Appendix~\ref{app:tau} for full results.

\begin{figure}[!t]
    \centering
    \includegraphics[width=\columnwidth]{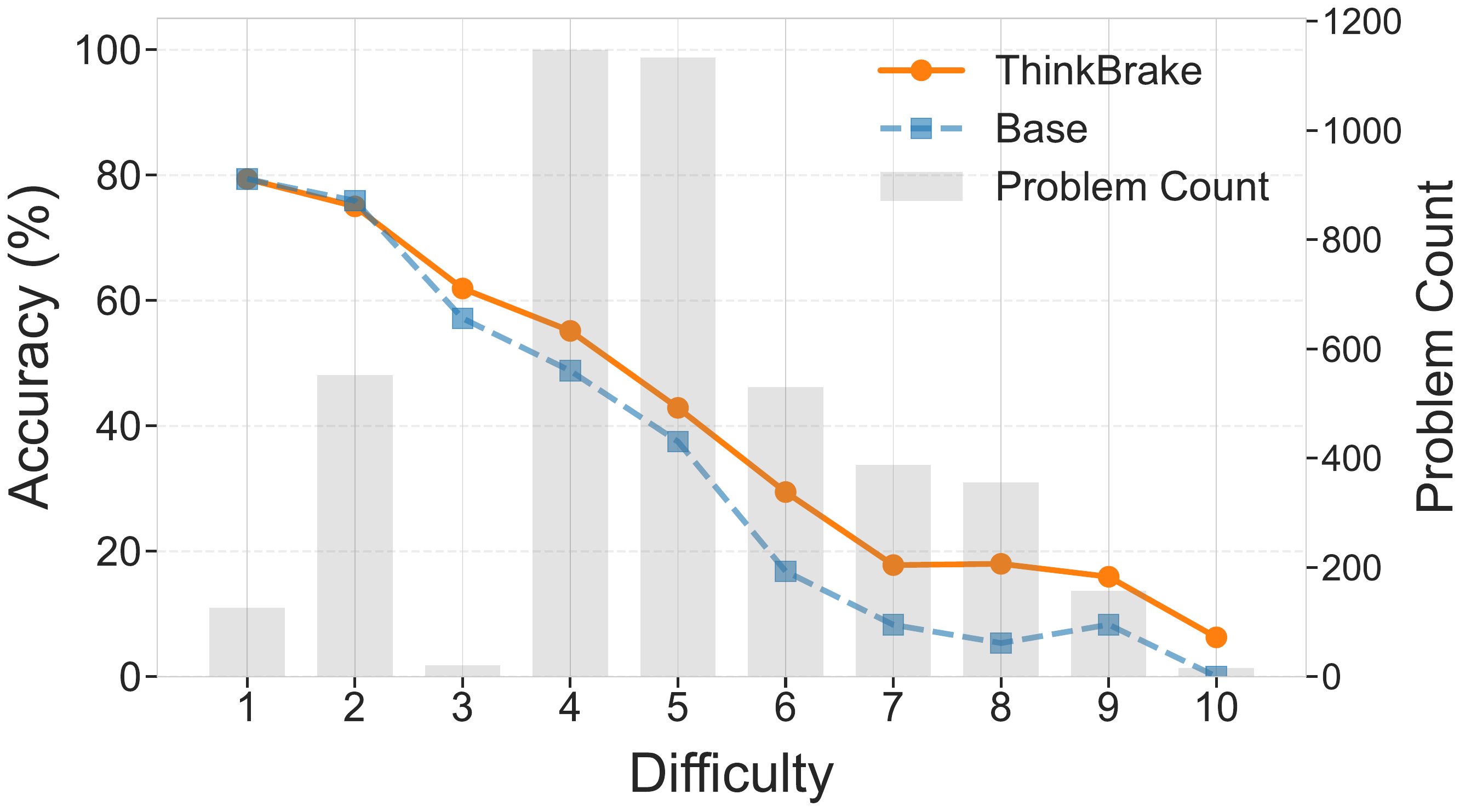}
    \caption{Accuracy by difficulty on Omni-MATH. \method{} maintains uniform performance across all levels. Grey bars show problem counts.}
    \label{fig:diff}
    \vspace{-10pt}
\end{figure}

\paragraph{Performance Across Problem Difficulties.}
\label{sec:difficulty_analysis}
We investigate whether \method{} trades performance between easy and hard problems, using the Omni-MATH dataset \citep{gao2024omni}, which provides difficulty labels from 1 to 10. Figure~\ref{fig:diff} shows \method{} maintains or improves accuracy uniformly across all difficulty levels without any trade-off. Notably, \method{} shows greater improvements on harder problems, suggesting that overthinking is more prevalent in challenging questions. We hypothesize that LRMs, uncertain about harder problems, continue generating spurious verification steps even after reaching a correct answer. This validates that efficiency gains come from preventing overthinking rather than sacrificing performance on hard problems. To further quantify how often early stopping changes correctness, we compute prediction transition matrices before and after \method{} (see Appendix~\ref{app:transition}). Across models, most examples remain unchanged (correct→correct or incorrect→incorrect), indicating \method{} primarily reduces computation without substantial accuracy trade-offs.


\subsection{Training from \method{} Data}
\label{sec:training}

Since \method{} generates concise trajectories that maintain or improve accuracy, we investigate whether this data can be used to train LRMs for efficient reasoning. We apply Direct Preference Optimization (DPO; \citealp{rafailov2023direct}) to Qwen3-4B-Thinking, treating \method{} trajectories as preferred and baseline trajectories as rejected. We construct 1.3K preference pairs from Omni-MATH problems. The training is highly efficient as it requires only 20 minutes on 2$\times$H200 GPUs (see Appendix~\ref{app:train_settings} for details).

Table~\ref{tab:qwen3_thinking_training} shows that DPO training successfully transfers \method{}'s efficiency to the LRM. Despite training only on mathematical reasoning, the model achieves 9--28\% thinking token reduction across all benchmarks while maintaining accuracy. Notably, improvements transfer to harder math problems (AIME), and to entirely out-of-domain tasks including GPQA-D, ARC-C, and tool usage. This demonstrates that the model learns general concise reasoning patterns rather than memorizing task-specific solutions, enabling efficient training of LRMs with minimal data and compute.

\begin{table}[!t]
\centering
\small
\begin{tabular}{lccc}
\toprule
Benchmark & Base & DPO & $\Delta$Tok \\
\midrule
GSM8K & 95.1 & 95.2 & -23.6\% \\
MATH500 & 95.3 & 96.7 & -27.7\% \\
AIME2024 & 77.1 & 80.7 & -8.8\% \\
AIME2025 & 76.2 & 76.0 & -9.5\% \\
GPQA-D & 64.1 & 61.5 & -16.2\% \\
ARC-C & 94.3 & 93.8 & -20.8\% \\ 
\midrule
BFCL-v1 & 88.3 & 88.0 & -16.6\% \\
BFCL-v2 & 82.5 & 83.6 & -10.5\% \\
Meta-Tool & 75.0 & 76.1 & -23.9\% \\
\bottomrule
\end{tabular}
\caption{Benchmark results for Qwen3-4B-Thinking trained with DPO on \method{}-generated data.}
\vspace{-10pt}
\label{tab:qwen3_thinking_training}
\end{table}

\section{Related Work}

\subsection{Concise Reasoning}
As LRMs often generate excessive tokens, recent work has studied efficient reasoning methods.

\paragraph{Test-Time Methods.} 
Several training-free approaches have been proposed, including sampling answers at each step using consistency as an early-stop signal \citep{mao2025early, liu-wang-2025-answer, wan2025reasoning}, employing external verifiers \citep{jiang2025flashthink}, or leveraging model representations via probing \citep{fu2025reasoning} and signals like confidence \citep{yang2025dynamic} and entropy \citep{laaouach2025haltcot}. Other heuristics include suppressing reflection tokens (e.g., ``Wait'', ``Hmm'') \citep{wang2025wait} or skipping explicit reasoning entirely \citep{li2025thinkless}. While effective in some cases, these methods typically rely on hand-crafted rules or require additional computational overhead from multiple rollouts or external models.

\paragraph{Training Methods.} 
An alternative approach trains models to produce concise traces directly. \citet{kang2025c3ot} compress long CoT and condition models to generate shorter reasoning, while subsequent work \citep{fang2025thinkless, song2025walk, fatemi2025concise, aggarwal2025l} presents RL frameworks for concise reasoning. However, these require significant computational resources and can potentially destabilize general model performance.

\subsection{Logit-Based Test-Time Realignment}
A complementary literature explores test-time realignment without retraining by applying logit-space shifts to the base model. Unlike heuristic test-time methods, these approaches implement principled objectives—such as RLHF alignment—directly at decoding time via logit manipulation. One line of work constructs such shifts using signals from aligned and base models \citep{mitchell2024an, liu2024tuning}, with controllable strength at decoding time \citep{liu2024decodingtime}, while related logit interventions can elicit specific abilities like reasoning \citep{zhang2025logit}. Another line leverages decoding-time steering for user-specific objectives, including modular rewards \citep{huang2025deal} or expert combinations \citep{liu2021dexperts}. Overall, these logit-based methods provide a theoretically grounded and flexible mechanism for test-time behavior control. \method{} falls within this framework: we implement a KL-regularized realignment objective that assigns a reward bonus to \texttt{</think>}, enabling principled concise reasoning.

\section{Conclusion}
We study overthinking in Large Reasoning Models across mathematical reasoning, scientific QA, and tool usage. Oracle experiments reveal that LRMs often reach correct intermediate solutions but fail to terminate, leaving substantial headroom for improved accuracy and efficiency. We introduce \method{}, a training-free decoding rule that triggers \texttt{</think>} based on log-probability margins at sentence boundaries. \method{} achieves favorable accuracy-efficiency trade-offs across multiple model families and benchmarks, is theoretically grounded as KL-regularized test-time realignment, and can train more efficient models via preference optimization.


\section*{Limitations}
\method{} comes with several limitations. First, \method{} requires explicit reasoning delimiters (\texttt{<think>} and \texttt{</think>}) and access to model logits, which may not be available in models with hidden chain-of-thought or black-box API settings. Second, the method introduces a hyperparameter $\tau$ that, while robust across a range of values, may require tuning in some settings. 
Finally, \method{} uses a local stopping criterion based on probability margins and does not explicitly reason about global answer correctness, which may lead to early termination when the model is confidently wrong.

\section*{Acknowledgements}
This work was supported by the National Research Foundation of Korea (NRF) grant (RS-2024-00333484) and by the Institute of Information \& Communications Technology Planning \& Evaluation (IITP) grant (RS-2025-02215122, Development and Demonstration of Lightweight AI Model for Smart Homes), all funded by the Korean government (MSIT). This research was also supported by the Korea Institute of Science and Technology Information (KISTI) in 2026 (No. (KISTI)K26L3M1C1), aimed at developing KONI (KISTI Open Neural Intelligence), a large language model specialized in science and technology.

\bibliography{custom}


\appendix
\clearpage

\section{Usage of Large Language Models}
Large language models were used for literature search, coding assistance, and proofreading only. They were not used for ideation, results, or analysis; all contributions and conclusions are the authors'.

\section{Example of Overthinking}
\label{app:overthinking_example}
Tables~\ref{tab:appendix-geometry} and~\ref{tab:trace-colored} present a BFCL example and the corresponding LRM response. Green highlights mark points at which terminating would yield a correct answer, whereas red highlights indicate an incorrect one. Notably, the model’s reasoning turns red after a certain point, illustrating overthinking.


\section{Experiment Details}
\label{app:exp_settings}
We use NVIDIA H200 GPU for all inference and training, with a fixed seed of 42.

\subsection{Model Details}
\label{app:model_id}
The official Huggingface names for the models we used are as follows in Table~\ref{tab:model_names}.

\begin{table}[h]
\centering
\small
\begin{tabular}{ll}
\toprule
\textbf{Our Name} & \textbf{Hf Model Id} \\
\midrule
Qwen3-4B-Thinking & \texttt{Qwen/Qwen3-4B-Thinking} \\ & \texttt{-2507} \\
Qwen3-4B & \texttt{Qwen/Qwen3-4B} \\
Qwen3-14B & \texttt{Qwen/Qwen3-14B} \\
Qwen3-32B & \texttt{Qwen/Qwen3-32B} \\
DeepSeek-R1-7B & \texttt{deepseek-ai/DeepSeek-} \\
& \texttt{R1-Distill-Qwen-7B} \\
Phi-4-Reasoning & \texttt{microsoft/Phi-4-reasoning} \\
Qwen3-Next-80B & \texttt{Qwen/Qwen3-Next-80B} \\ &\texttt{-A3B-Thinking} \\
\bottomrule
\end{tabular}
\caption{Model names and Huggingface identifiers.}
\label{tab:model_names}
\end{table}

\subsection{Inference Details}
For all inference tasks, we use the following hyperparameters following the official guidelines for each model \citep{yang2025qwen3,abdin2025phi,guo2025deepseek}. For Qwen3-4B-Thinking, we extend the reasoning budget to 32k for AIME benchmarks only.
\begin{table}[h]
\centering
\resizebox{0.8\columnwidth}{!}{%
\begin{tabular}{lccc}
\toprule
\textbf{Parameter} & \textbf{DeepSeek} & \textbf{Qwen3} & \textbf{Phi-4} \\
\midrule
\method{}-$\tau$ & 0.1 & 0.1 & 2.5 \\
\midrule
\multicolumn{4}{l}{\textit{Sampling Parameters}} \\
\midrule
Temperature & 0.6 & 0.6 & 0.8 \\
Top-p & 1 & 0.95 & 0.95 \\
Top-k & - & 20 & 50 \\
\midrule
\multicolumn{4}{l}{\textit{Token Budgets}} \\
\midrule
Reasoning Budget & 16384 & 16384 & 16384 \\
Answer Budget & 4096 & 4096 & 4096 \\
\bottomrule
\end{tabular}%
}
\caption{Inference hyperparameters for different models.}
\label{tab:inference_hyperparams}
\end{table}

\subsection{Training on \method{}}
\label{app:train_settings}
This section describes the experimental details for \S\ref{sec:experiments}. To curate a training dataset, we sampled $\sim$1.3K examples from Omni-MATH, with the preferred responses generated using \method{} and dispreferred responses generated without \method{}. We applied the following filtering criteria: (1) when both methods produce correct answers, we only include examples where the baseline token length exceeds the \method{} token length, demonstrating efficiency gains; (2) when \method{} produces correct answers but the baseline fails, we include all such examples; (3) all other cases are excluded from the training set. This curation process ensures that our training data emphasizes both correctness improvements and computational efficiency.

We performed Direct Preference Optimization (DPO; \citealp{rafailov2023direct}) using the hyperparameters listed in Table~\ref{tab:training_hyperparams}. Training was conducted with LoRA~\citep{hu2022lora} for parameter-efficient fine-tuning in around 20 minutes of wall-clock time.

\begin{table}[h]
\centering
\begin{tabular}{lc}
\toprule
\textbf{Hyperparameter} & \textbf{Value} \\
\midrule
Learning Rate & $2 \times 10^{-5}$ \\
Batch Size (per device) & 8 \\
Gradient Accumulation Steps & 4 \\
Effective Batch Size & 64 \\
Number of GPUs & 2 $\times$ H200 \\
Training Epochs & 10 \\
LoRA Rank ($r$) & 32 \\
LoRA Alpha ($\alpha$) & 32 \\
Training Samples & 1,348 \\
\bottomrule
\end{tabular}
\caption{Training hyperparameters for DPO.}
\label{tab:training_hyperparams}
\end{table}

\subsection{System Prompts}
\label{app:system_prompts}
Following the official guides \citep{abdin2025phi,guo2025deepseek,yang2025qwen3} we use the following prompts in Figures~\ref{fig:qwen3_math}, \ref{fig:qwen3_general},~\ref{fig:deepseek_math} and~\ref{fig:phi4_system} for our experiments.

\clearpage
\subsection{Baseline Methods}
\label{app:baseline}
This section describes the details on the baseline methods: \\[1em]
\textbf{NoWait}~\citep{wang2025wait}: removes delay/filler tokens (e.g., “wait”, “hmm”) during decoding to stop overthinking and allow for earlier termination. We use the following list in Table~\ref{tab:suppress-keywords} for NoWait.
\begin{table}[htbp]
\centering
\small
\begin{tabular}{p{\linewidth}}
\toprule
\textbf{Keywords} \\
\midrule
wait, hmm, hmmm, but, however, check, verify, alternatively \\
\bottomrule
\end{tabular}
\caption{Keyword list for NoWait.}
\label{tab:suppress-keywords}
\end{table}
\\
\textbf{ThinkLess}~\citep{li2025thinkless}: forces immediate termination by emitting \texttt{</think>} right after the initial \texttt{<think>}, minimizing the thinking process to two special tokens: \texttt{<think>}\texttt{</think>}.

\paragraph{Dynasor-CoT}~\citep{fu2025reasoning}: injects probing prompts at regular intervals to extract intermediate answers, terminating reasoning early if those answers demonstrate consistency across consecutive steps.
\paragraph{DEER}~\citep{yang2025dynamic}: monitors the model's reasoning for linguistic transition markers (e.g., `Wait', `Alternatively'), temporarily interrupts the process at these points to induce a trial answer, and evaluates the model's internal probability confidence in that answer—exiting to output it if the confidence threshold is met, or discarding it and resuming the chain of thought if it is not.

\subsection{Hyperparameter Selection}
\label{app:hyperparameter_selection}
\begin{figure}[H]
    \centering
    \includegraphics[width=\columnwidth]{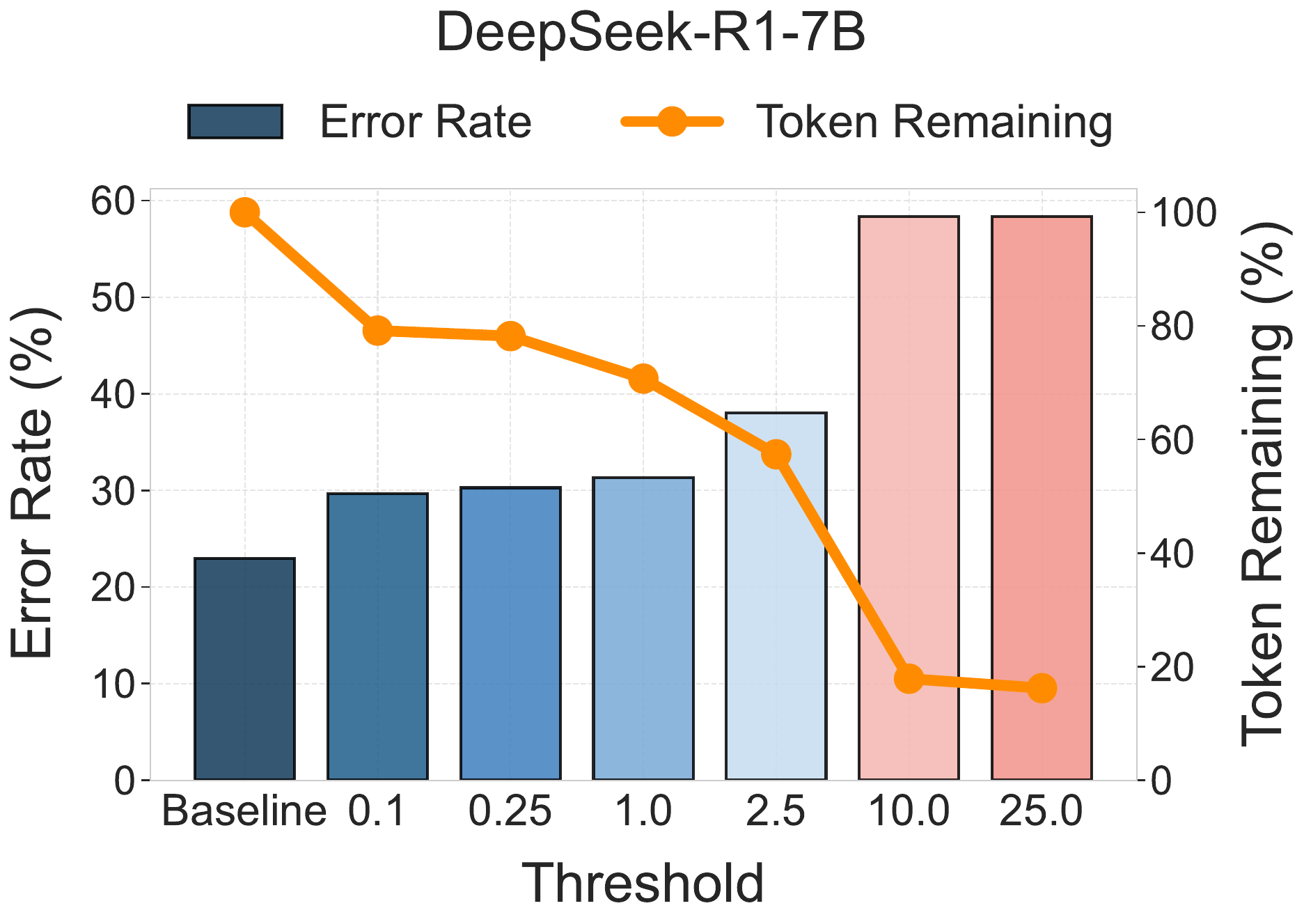}
    \caption{$\tau$ search from error rate and thinking token usage for \method{} on DeepSeek-R1-7B.}
    \label{fig:tau_select1}
\end{figure}

\begin{figure}[H]
    \centering
    \includegraphics[width=\columnwidth]{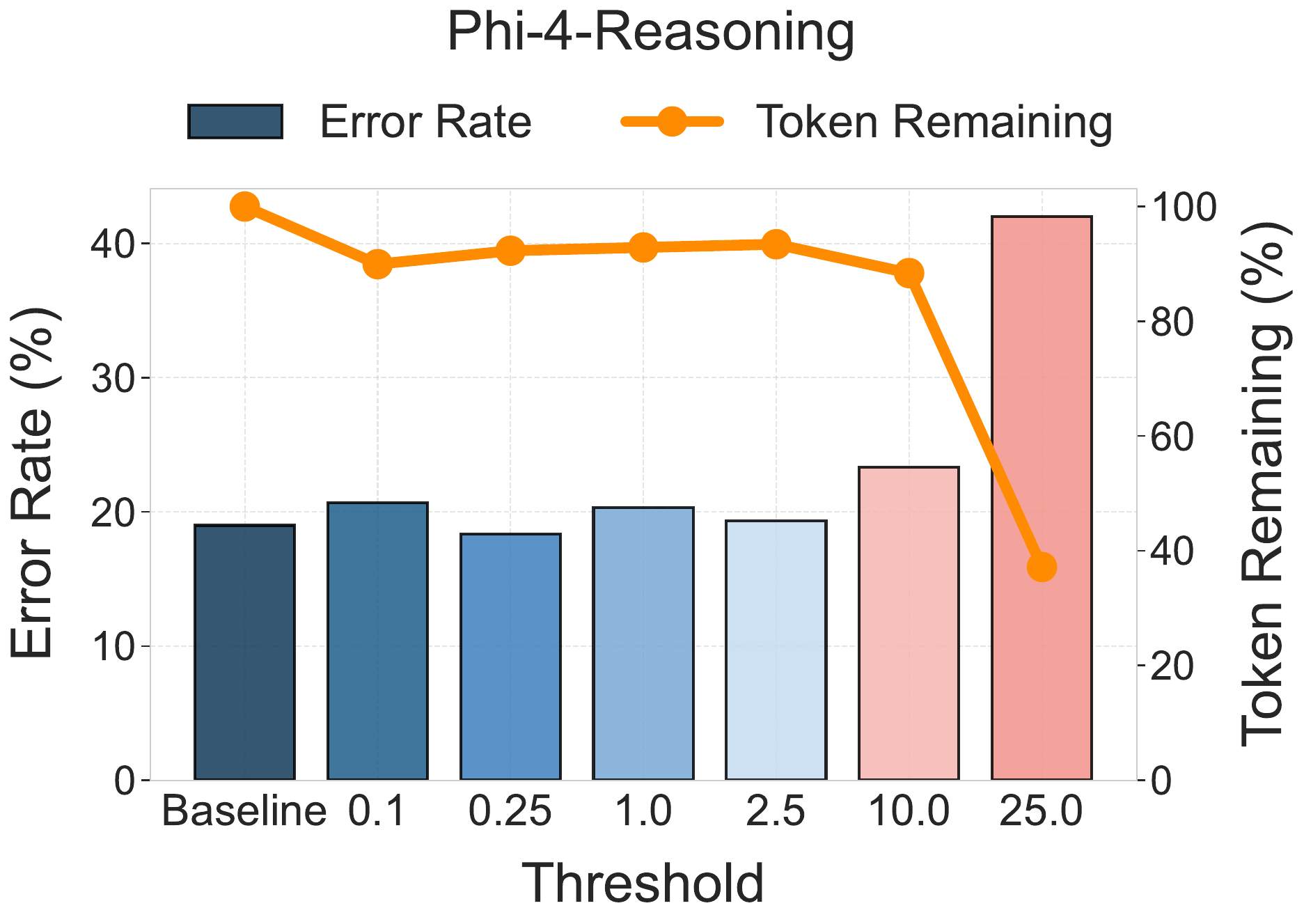}
    \caption{$\tau$ search from error rate and thinking token usage for \method{} on Phi-4-Reasoning.}
    \label{fig:tau_select2}
\end{figure}

Figures~\ref{fig:tau_select1} and~\ref{fig:tau_select2} show the hyperparameter search for additional models.
 
\section{Why Log-Space Margins Are Robust}
\label{app:log_explanation}
\begin{figure}[htbp]
    \centering
    \includegraphics[width=\columnwidth]{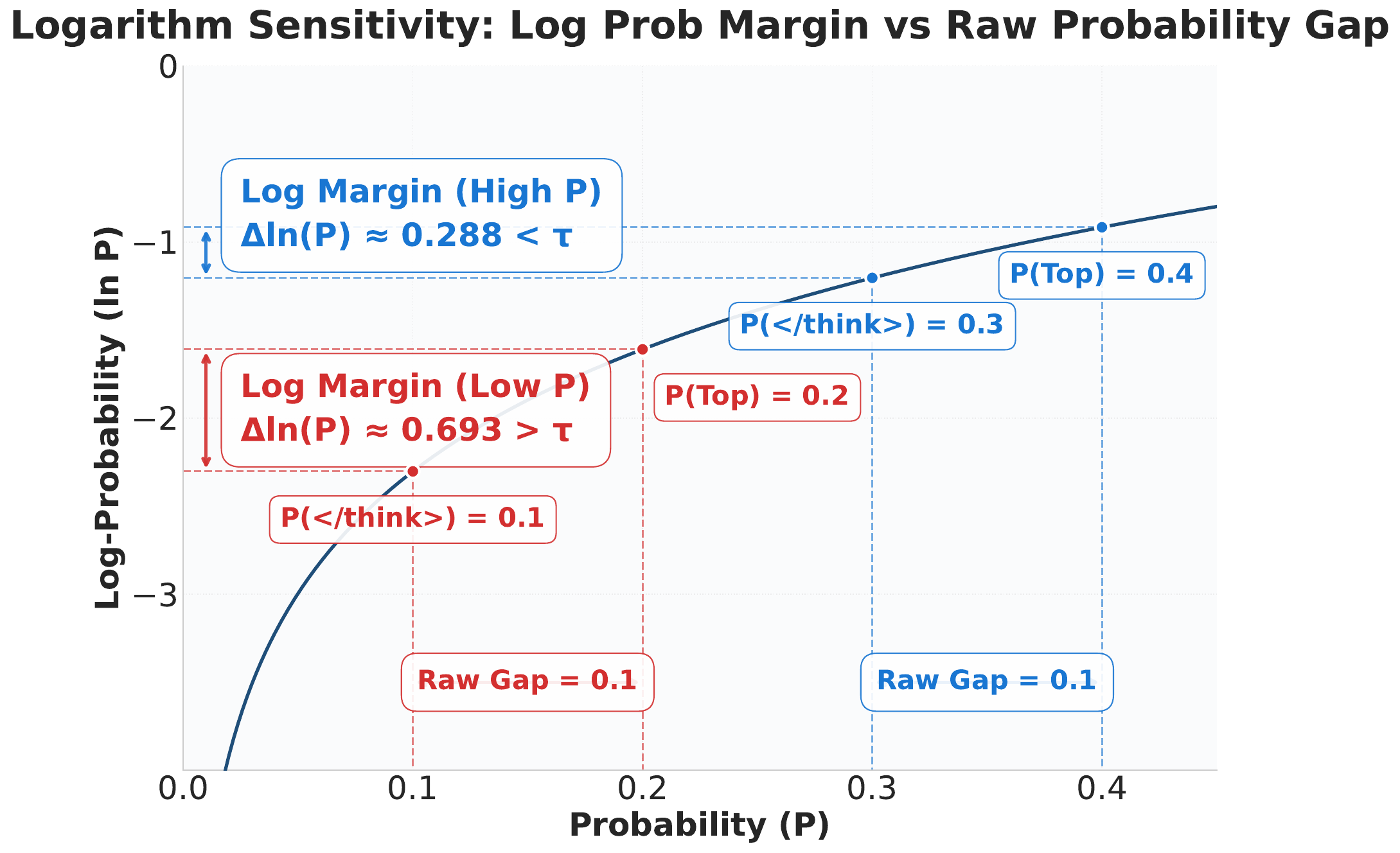}
    \caption{Log-probability plot with two illustrative cases, $\Delta$P1 and $\Delta$P2. Here, $\Delta$P1 denotes a scenario where the most likely token has high probability (0.4), and $\Delta$P2 denotes a scenario where the most likely token has lower probability (0.2), while the raw gap is the same (i.e., $\Delta$P1  = $\Delta$P2).}
    \label{fig:log}
\end{figure}

As shown in Tables~\ref{tab:math_results_full}, ~\ref{tab:bfcl_meta_results_full} and \S\ref{sec:results_logprob}, linear probability gaps are insufficient for \method{}. A meaningful log-probability gap only arises when both competing probabilities are relatively high (see Figure~\ref{fig:log}), indicating that the model is genuinely confident at that step. If we used a linear probability gap, both $\Delta$P1 and $\Delta$P2 would trigger early termination; in contrast, using a log-probability gap triggers only for $\Delta$P1—where both the top token and \texttt{</think>} have high probabilities—signaling confidence in stopping the reasoning.

\setlength{\textfloatsep}{5pt} 

\paragraph{Entropy State.}\label{app:entropy} Table~\ref{tab:entropy_results} shows the ratio of \texttt{</think>} in the top 20 tokens and median entropy for the log-probability setting (\method{}) and the raw-probability variant (\method{}\nobreakdash-p). \method{} terminates almost exclusively in low-entropy states and only when \texttt{</think>} ranks within the top 20 tokens, while the raw-probability variant (\method{}-p) frequently triggers in high-entropy states even when \texttt{</think>} is not in the top 20, confirming that the log design is robust precisely where the raw-probability design is more sensitive.

\begin{table}[htbp]
\centering
\resizebox{\columnwidth}{!}{
\begin{tabular}{llrc}
\toprule
Category & Method & Ratio in Top-20 & Median Entropy \\
\midrule
\multirow{2}{*}{Math} & \method{} & 99.9\% & 0.018 \\
 & \method{}-p & 66.6\% & 0.264 \\
\midrule
\multirow{2}{*}{Science} & \method{} & 100.0\% & 0.036 \\
 & \method{}-p & 65.3\% & 0.458 \\
\bottomrule
\end{tabular}
}
\caption{Entropy and top-20 ratio results on Qwen3-4B-Thinking across math and science benchmarks.}
\label{tab:entropy_results}
\end{table}

\section{Extensive Results}
\label{app:more_results}
In this section, we provide extensive results across all benchmarks, models, and hyperparameter combinations.

\paragraph{Token Count.} Table~\ref{tab:base_token_counts_split} shows the average token count per base model and benchmark. We consistently observe that difficult tasks such as AIME and GPQA-D require larger token budgets.

\paragraph{Full Tables.} 
\label{app:full} Tables~\ref{tab:math_results_full} and~\ref{tab:bfcl_meta_results_full} present extensive results from the experiments in \S\ref{sec:experiments}. Specifically, Table~\ref{tab:math_results_full} shows the full results including all baseline methods across all models. Table~\ref{tab:bfcl_meta_results_full} shows the complete results for BFCL-v1, BFCL-v2 (including Simple and Multiple function categories), and Meta-Tool benchmarks. We observe that \method{} achieves a better performance-token reduction tradeoff overall. Furthermore, Table~\ref{tab:variance_ci_results} reports the variances and 95\% confidence intervals (CIs) for accuracy and token counts on the AIME benchmarks, computed over 32 independent runs. Notably, \method{}'s token consumption falls entirely below the baseline's CI, confirming a statistically meaningful reduction in reasoning length.

\paragraph{Token Reduction and Model Size.}
\label{app:80B} Table~\ref{tab:reasoning_trace_trend_results} shows that larger models tend to produce shorter reasoning traces and there is less redundant thinking to prune. Qwen3-Next-80B, despite being larger, produces substantially longer traces than Qwen3-32B, and correspondingly yields larger token reductions. We suggest that token reduction rate depends on the reasoning trace length rather than model size.

\paragraph{Premature Exiting and Spurious Reasoning.}\label{app:spurious}
To investigate the performance degradation associated with \method{}, we analyzed the per-problem token counts of \method{} versus an oracle setting on the highly challenging AIME benchmarks. Table~\ref{tab:oracle_thinkbrake_token_delta} reveals that among the 16 instances where \method{} failed but the oracle succeeded, \method{} exited prior to the oracle in only 2 cases. This indicates that premature termination is not the primary reason for incorrect answers. Instead, errors appear to stem from subsequent spurious reasoning that misleads the LRMs. A detailed example is shown in Table~\ref{tab:spurious_trace}, where the model successfully solves the problem but then engages in flawed reasoning before \method{} finally terminates generation. We hypothesize that highly verbose models, such as Qwen3-4B-Thinking (which produces the longest average reasoning traces), are especially vulnerable to this type of reasoning degradation.

\begin{figure*}[t]
    \centering
    \includegraphics[width=\textwidth]{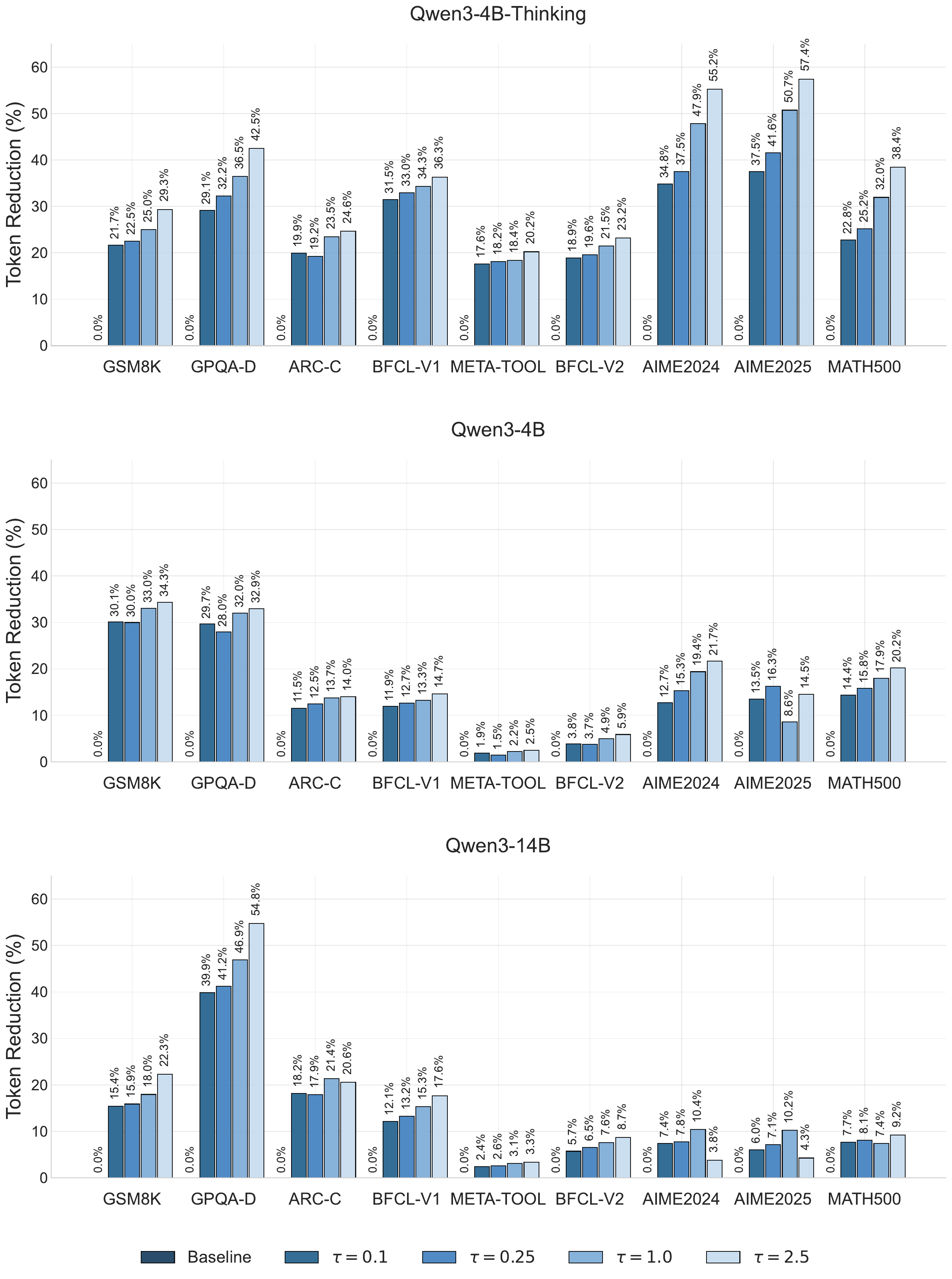}
  \caption{Token reduction across various threshold values $\tau$ for Qwen3-4B, Qwen3-4B-Thinking, and Qwen3-14B.}
  \label{fig:token_reduction_qwen_small}
\end{figure*}

\begin{figure*}[t]
    \centering
    \includegraphics[width=\textwidth]{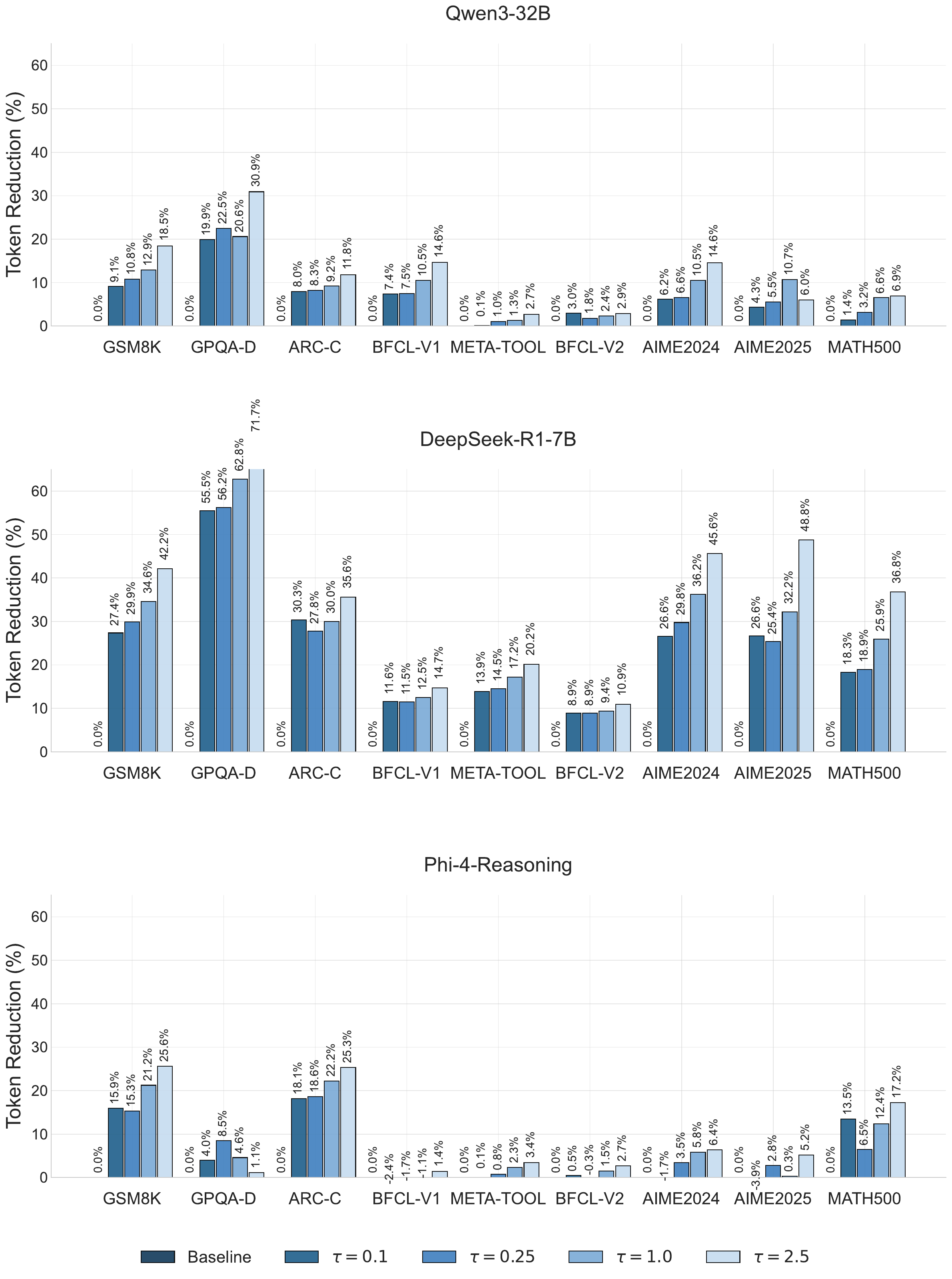}
  \caption{Token reduction across various threshold values $\tau$ for Qwen3-32B, DeepSeek-R1-7B, and Phi-4-Reasoning.}
  \label{fig:token_reduction_others}
\end{figure*}

\begin{figure*}[t]
    \centering
    \includegraphics[width=\textwidth]{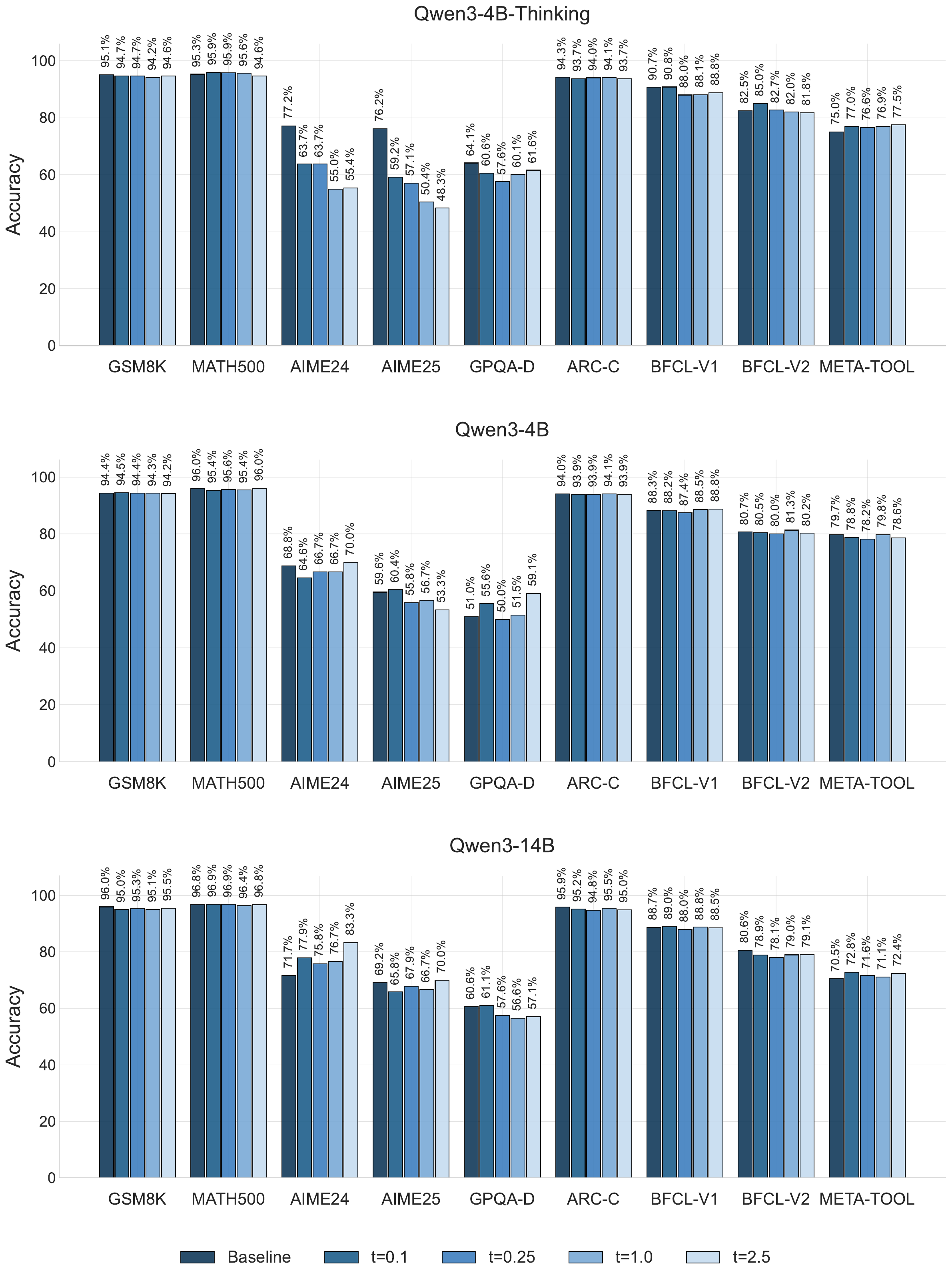}
  \caption{Accuracy comparison across various threshold values $\tau$ for Qwen3-4B-Thinking, Qwen3-4B, and Qwen3-14B.}
  \label{fig:accuracy_qwen_small}
\end{figure*}

\begin{figure*}[t]
    \centering
    \includegraphics[width=\textwidth]{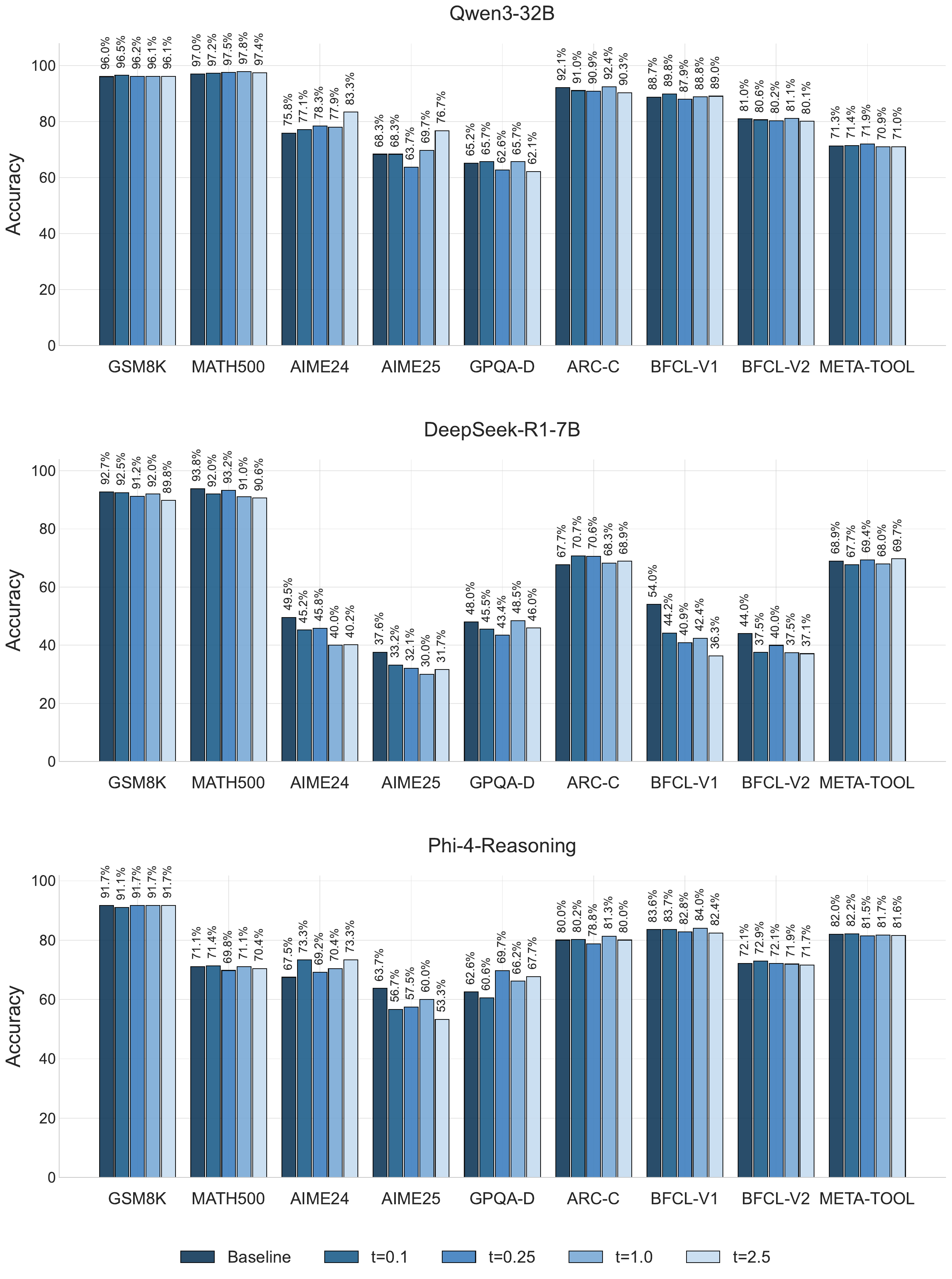}
  \caption{Accuracy comparison across various threshold values $\tau$ for Qwen3-32B, DeepSeek-R1-7B, and Phi-4-Reasoning.}
  \label{fig:accuracy_others}
\end{figure*}

\paragraph{$\tau$ Ablation.}
\label{app:tau} Figures~\ref{fig:token_reduction_qwen_small},~\ref{fig:token_reduction_others},~\ref{fig:accuracy_qwen_small}, and~\ref{fig:accuracy_others} show extensive results across various threshold values $\tau \in \{0.1, 0.25, 1.0, 2.5\}$. The first two figures display token reduction percentage compared to the base model, while the latter two show accuracy comparison across various $\tau$ values. We observe that larger $\tau$ values result in greater token reduction, and accuracy is maintained across various $\tau$ settings.

\begin{figure*}[t]
    \includegraphics[width=\textwidth]{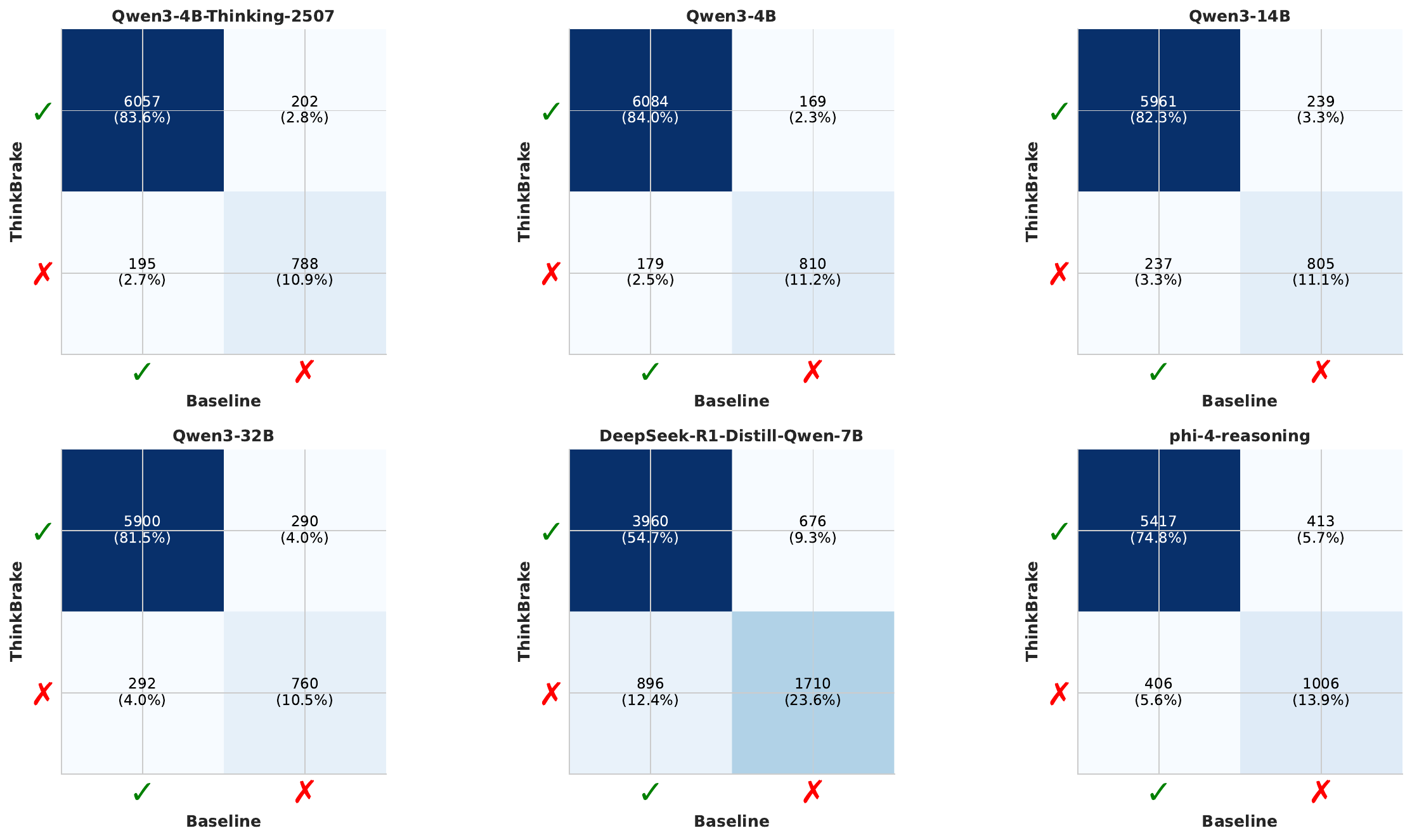}\\[1em]
  \caption{Transition matrices showing prediction transitions before and after applying \method{} for all models, aggregated across all benchmarks. High diagonal values indicate preserved accuracy.}
  \label{fig:confusion_matrix}
\end{figure*}

\paragraph{Transition Matrix.}
\label{app:transition} Figure~\ref{fig:confusion_matrix} shows the transition matrices for all models (aggregated across all benchmarks), illustrating model behavior before and after applying \method{}. The matrices show high values along the diagonal (top-left and bottom-right), indicating that \method{} preserves accuracy without introducing significant changes to correct or incorrect predictions.

\clearpage

\clearpage
\begin{table*}[htbp]
\centering
\begin{tabular}{p{2\columnwidth}}
\toprule
\textbf{Question} \\ \midrule
Find the area and perimeter of a circle with a radius of 5 and also find the circumference of a circle with diameter of 10. \\ \midrule

\textbf{Prompt} \\ \midrule
\texttt{system:} \\
\# Tools \\
You may call one or more functions to assist with the user query. \\
You are provided with function signatures within \textless tools\textgreater{} XML tags: \\

\textless tools\textgreater{} \\
\{ "name": "circle.calculate\_circumference", \\
\hspace*{0.5cm} "description": "Calculate the circumference of a circle based on the diameter.", \\
\hspace*{0.5cm} "parameters": \{ \\
\hspace*{1cm} "type": "dict", \\
\hspace*{1cm} "properties": \{ \\
\hspace*{1.5cm} "diameter": \{ "type": "integer", "description": "The diameter of the circle." \} \\
\hspace*{1cm} \}, \\
\hspace*{1cm} "required": ["diameter"] \\
\hspace*{0.5cm} \} \\
\}, \\
\{ "name": "circle.calculate\_area", \\
\hspace*{0.5cm} "description": "Calculate the area of a circle based on the radius.", \\
\hspace*{0.5cm} "parameters": \{ \\
\hspace*{1cm} "type": "dict", \\
\hspace*{1cm} "properties": \{ \\
\hspace*{1.5cm} "radius": \{ "type": "integer", "description": "The radius of the circle." \} \\
\hspace*{1cm} \}, \\
\hspace*{1cm} "required": ["radius"] \\
\hspace*{0.5cm} \} \\
\}, \\
\{ "name": "rectangle.calculate\_perimeter", \\
\hspace*{0.5cm} "description": "Calculate the perimeter of a rectangle based on the length and breadth.", \\
\hspace*{0.5cm} "parameters": \{ \\
\hspace*{1cm} "type": "dict", \\
\hspace*{1cm} "properties": \{ \\
\hspace*{1.5cm} "length": \{ "type": "integer", "description": "The length of the rectangle." \}, \\
\hspace*{1.5cm} "breadth": \{ "type": "integer", "description": "The breadth of the rectangle." \} \\
\hspace*{1cm} \}, \\
\hspace*{1cm} "required": ["length", "breadth"] \\
\hspace*{0.5cm} \} \\
\} \\
\textless/tools\textgreater{} \\

For each function call, return a json object with function name and arguments within \textless tool\_call\textgreater{} XML tags: \\
\textless tool\_call\textgreater{} \\
\{ "name": \textless function-name\textgreater{}, "arguments": \textless args-json-object\textgreater{} \} \\
\textless/tool\_call\textgreater{} \\

\texttt{user:} The above geometry question. \\
\bottomrule
\end{tabular}
\caption{Full question and prompt.}
\label{tab:appendix-geometry}
\end{table*}

\clearpage
\onecolumn
\begin{longtable}{p{\textwidth}}
\toprule
\small\ttfamily
\endfirsthead

\toprule
\small\ttfamily
\endhead

\bottomrule
\endfoot

\endlastfoot

\hlg{<think>}\\
\hlg{Okay, let's tackle this problem. The user wants the area and perimeter of a circle with radius 5, and the circumference of a circle with diameter 10.}\\
\hlg{First, I need to check which tools are available.}\\
\hlg{There's circle.calculate\_area which takes radius, and rectangle.calculate\_perimeter but wait, the user is asking for the perimeter of a circle? Wait, no.}\\
\hlg{Wait, the problem says "area and perimeter of a circle".}\\
\hlg{Wait, circles don't have perimeters; they have circumferences.}\\
\hlr{Maybe the user made a mistake here.}\\
\hlg{Let me read again.}\\
\hlg{The user says: "Find the area and perimeter of a circle with a radius of 5 and also find the circumference of a circle with diameter of 10."}\\
\hlg{Hmm, perimeter for a circle is called circumference.}\\
\hlg{So maybe the user is confused.}\\
\hlg{But according to the tools provided, there's circle.calculate\_circumference (for diameter), circle.calculate\_area (for radius), and rectangle.calculate\_perimeter (for length and breadth).}\\
\hlg{Wait, the user is asking for perimeter of a circle.}\\
\hlg{But the tools don't have a circle perimeter function.}\\
\hlg{Wait, the rectangle function is for perimeter of a rectangle, which is different.}\\
\hlg{So maybe the user intended to say circumference for the circle's perimeter.}\\
\centerline{\vdots}\\
\hlr{Because perimeter is a term for 2D shapes, but for circles it's circumference.}\\
\hlr{Assuming that the user wants:}\\
\hlr{- Area of radius 5 circle: call calculate\_area with radius=5.}\\
\hlr{- Circumference of radius 5 circle (which they called perimeter): but since the tool requires diameter, we can compute diameter as 10 (since radius is 5), so call calculate\_circumference with diameter=10.}\\
\hlr{- Also, circumference with diameter 10: call calculate\_circumference with diameter=10.}\\
\centerline{\vdots}\\
\hlr{So three tool calls? Or is the perimeter part a mistake and the user just meant circumference?}\\
\hlr{Wait, the rectangle.calculate\_perimeter is for rectangles, so that's not relevant here.}\\
\hlr{The user is talking about circles, so we ignore the rectangle tool.}\\
\hlr{So, the correct tool calls are:}\\
\hlr{1. circle.calculate\_area(radius=5)}\\
\hlr{2. circle.calculate\_circumference(diameter=10)}\\
\hlr{3. circle.calculate\_circumference(diameter=10)}\\
\hlr{But in reality, the user is asking for two separate values: perimeter (which is circumference) for radius 5 (diameter 10) and circumference for diameter 10.}\\
\hlr{So they are the same calculation, but the user is asking for it twice.}\\
\centerline{\vdots}\\
\hlr{- For the perimeter of radius 5 circle: since perimeter of a circle is circumference, which is calculated via diameter.}\\
\hlr{So diameter is .}\\
\bottomrule

\caption{Qwen3-4B thinking trajectory.}\label{tab:trace-colored}
\end{longtable}
\clearpage
\twocolumn

\begin{figure}[htbp]
\centering
\begin{tcolorbox}[title=Qwen3 Mathematical Reasoning Prompt, halign=left, boxrule=0.5pt]\normalsize
\texttt{<|im\_start|>user}

\{\{ problem \}\}

Please reason step by step, and put your final answer within \textbackslash boxed\{\}.

\texttt{<|im\_end|>}

\texttt{<|im\_start|>assistant}

\texttt{<think>}

\{\{ reasoning process \}\}

\texttt{</think>}

\{\{ final answer \}\}

\texttt{<|im\_end|>}
\end{tcolorbox}
\caption{Prompt template for Qwen3 on mathematical reasoning tasks. Similar to DeepSeek, the model provides reasoning within think tags before the final boxed answer.}
\label{fig:qwen3_math}
\end{figure}

\begin{figure}[htbp]
\centering
\begin{tcolorbox}[title=Qwen3 Multiple-Choice Reasoning Prompt, halign=left, boxrule=0.5pt]\normalsize
\texttt{<|im\_start|>user}

\{\{ problem \}\}

Please show your choice in the answer field with only the choice letter, e.g., "answer": "C".

\texttt{<|im\_end|>}

\texttt{<|im\_start|>assistant}

\texttt{<think>}

\{\{ reasoning process \}\}

\texttt{</think>}

\{\{ final answer with choice letter \}\}

\texttt{<|im\_end|>}
\end{tcolorbox}
\caption{Prompt template for Qwen3 on general multiple-choice reasoning tasks. The model is instructed to format its final answer as a single choice letter.}
\label{fig:qwen3_general}
\end{figure}

\begin{figure}[htbp]
\centering
\begin{tcolorbox}[title=DeepSeek Reasoning Prompt, halign=left, boxrule=0.5pt]\normalsize
\texttt{<|begin\_of\_sentence|><|User|>}\{\{ problem \}\}

Please reason step by step, and put your final answer within \textbackslash boxed\{\}.

\texttt{<|Assistant|><think>}

\{\{ reasoning process \}\}

\texttt{</think>}

\{\{ final answer \}\}

\texttt{<|end\_of\_sentence|>}
\end{tcolorbox}
\caption{Prompt template for DeepSeek on mathematical reasoning tasks. The model is instructed to provide step-by-step reasoning within think tags, followed by the final answer in boxed notation.}
\label{fig:deepseek_math}
\end{figure}

\begin{figure}[htbp]
\centering
\begin{tcolorbox}[title=Phi-4-Reasoning System Prompt, halign=left, boxrule=0.5pt]\small
\texttt{<|im\_start|>system<|im\_sep|>}

You are Phi, a language model trained by Microsoft to help users. Your role as an assistant involves thoroughly exploring questions through a systematic thinking process before providing the final precise and accurate solutions.

This requires engaging in a comprehensive cycle of analysis, summarizing, exploration, reassessment, reflection, backtracing, and iteration to develop well-considered thinking process.

Please structure your response into two main sections: Thought and Solution using the specified format:

\texttt{<think> \{Thought section\} </think> \{Solution section\}}

\textbf{In the Thought section:} Detail your reasoning process in steps. Each step should include detailed considerations such as analysing questions, summarizing relevant findings, brainstorming new ideas, verifying the accuracy of the current steps, refining any errors, and revisiting previous steps.

\textbf{In the Solution section:} Based on various attempts, explorations, and reflections from the Thought section, systematically present the final solution that you deem correct. The Solution section should be logical, accurate, and concise and detail necessary steps needed to reach the conclusion.

Now, try to solve the following question through the above guidelines:

\texttt{<|im\_end|>}
\texttt{<|im\_start|>user<|im\_sep|>}
\texttt{\{\{ problem \}\}}
\texttt{<|im\_end|>}
\texttt{<|im\_start|>assistant<|im\_sep|><think>}
\texttt{\{\{ reasoning process \}\}}
\texttt{</think>}
\texttt{\{\{ final answer \}\}}
\texttt{<|im\_end|>}
\end{tcolorbox}
\caption{Prompt template for Phi-4-Reasoning with comprehensive system instructions. The model is guided to use a systematic thinking process with explicit thought and solution sections.}
\label{fig:phi4_system}
\end{figure}

\clearpage
\onecolumn
\begin{table*}[t]
\centering
\resizebox{\textwidth}{!}{
\begin{tabular}{lcccccccccc}
\toprule
\textbf{(a) Math} & \multicolumn{2}{c}{GSM8K} & \multicolumn{2}{c}{MATH500} & \multicolumn{2}{c}{AIME2024} & \multicolumn{2}{c}{AIME2025} & \multicolumn{2}{c}{Avg.} \\
\cmidrule(lr){2-3}\cmidrule(lr){4-5}\cmidrule(lr){6-7}\cmidrule(lr){8-9}\cmidrule(lr){10-11}
Method & Acc & $\Delta$Tok & Acc & $\Delta$Tok & Acc & $\Delta$Tok & Acc & $\Delta$Tok & Acc & $\Delta$Tok \\
\midrule
Base   & 95.1 & --      & 95.3 & --      & 77.1 & --      & 76.2 & --      & 86.0 & --      \\
Oracle & 97.8 & -69.1\% & 99.8 & -60.0\% & 86.7 & -41.6\% & 93.3 & -45.5\% & 94.4 & -54.1\% \\
\cmidrule(lr){1-11}
\textit{Avg. Base Token} & \multicolumn{2}{c}{\textit{1,578}} & \multicolumn{2}{c}{\textit{6,360}} & \multicolumn{2}{c}{\textit{19,294}} & \multicolumn{2}{c}{\textit{21,317}} & \multicolumn{2}{c}{\textit{12,137}} \\
\bottomrule
\end{tabular}%
}

\vspace{0.5em} 

\resizebox{\textwidth}{!}{
\begin{tabular}{lcccccccccc}
\toprule
\textbf{(b) Tool} & \multicolumn{2}{c}{simple} & \multicolumn{2}{c}{multiple} & \multicolumn{2}{c}{parallel} & \multicolumn{2}{c}{multi-parallel} & \multicolumn{2}{c}{Avg.} \\
\cmidrule(lr){2-3}\cmidrule(lr){4-5}\cmidrule(lr){6-7}\cmidrule(lr){8-9}\cmidrule(lr){10-11}
Method & Acc & $\Delta$Tok & Acc & $\Delta$Tok & Acc & $\Delta$Tok & Acc & $\Delta$Tok & Acc & $\Delta$Tok \\
\midrule
Base   & 88.2 & --      & 90.5 & --      & 93.5 & --      & 90.5 & --      & 90.7 & --      \\
Oracle & 91.5 & -81.3\% & 98.5 & -93.6\% & 96.0 & -89.7\% & 95.5 & -90.1\% & 95.4 & -88.7\% \\
\cmidrule(lr){1-11}
\textit{Avg. Base Token} & \multicolumn{2}{c}{\textit{1,155}} & \multicolumn{2}{c}{\textit{985}} & \multicolumn{2}{c}{\textit{1,200}} & \multicolumn{2}{c}{\textit{1,621}} & \multicolumn{2}{c}{\textit{1,240}} \\
\bottomrule
\end{tabular}%
}

\caption{Accuracy (\%) and $\Delta$Tok (token reduction, \%) on Math and Tool benchmarks. The results represent performance under oracle \texttt{</think>} stopping compared to the base Qwen3-4B-Thinking model.}
\label{tab:early_stop_split}
\end{table*}

\begin{table*}[t]
\centering
\footnotesize
\setlength{\tabcolsep}{7pt}
\renewcommand{\arraystretch}{1.2}
\resizebox{\textwidth}{!}{

\begin{tabular}{lcccccc}
\toprule
Model & GSM8K & MATH500 & AIME 2024 & AIME 2025 & GPQA-D & ARC-C \\
\midrule
Qwen3-4B-Thinking & 1,578 & 6,360 & 19,294 & 21,317 & 8,455 & 1,178 \\
Qwen3-4B & 2,407 & 5,250 & 13,000 & 15,163 & 7,497 & 798 \\
Qwen3-14B & 1,912 & 4,938 & 12,807 & 14,198 & 7,379 & 621 \\
Qwen3-32B & 1,794 & 4,650 & 12,278 & 13,727 & 6,626 & 726 \\
DeepSeek-R1-7B & 1,674 & 4,062 & 12,146 & 12,476 & 7,097 & 770 \\
Phi-4-Reasoning & 1,332 & 2,664 & 9,616 & 10,599 & 7,896 & 1,360 \\
Qwen3-Next-80B & 1,169 & 4,547 & 15,538 & 16,976 & 8,025 & 807\\
\bottomrule
\end{tabular}} \\[1em]
\setlength{\tabcolsep}{3.5pt}
\resizebox{\textwidth}{!}{
\begin{tabular}{lcccccccccc}
\toprule
& \multicolumn{4}{c}{BFCL-v1} & \multicolumn{4}{c}{BFCL-v2} & \multicolumn{2}{c}{Meta-Tool} \\
\cmidrule(lr){2-5} \cmidrule(lr){6-9} \cmidrule(lr){10-11}
Model & Simple & Multiple & Parallel & Multi-Par. & Simple & Multiple & Parallel & Multi-Par. & Single & Multiple \\
\midrule
Qwen3-4B-Thinking & 1,072 & 965 & 1,199 & 1,621 & 1,094 & 1,832 & 1,473 & 2,397 & 1,946 & 1,336 \\
Qwen3-4B & 607 & 744 & 844 & 1,046 & 653 & 1,426 & 837 & 1,555 & 1,069 & 856 \\
Qwen3-14B & 545 & 725 & 834 & 975 & 612 & 1,372 & 944 & 1,470 & 1,091 & 873 \\
Qwen3-32B & 553 & 750 & 826 & 991 & 617 & 1,342 & 725 & 1,572 & 1,048 & 857 \\
DeepSeek-R1-7B & 669 & 959 & 815 & 1,152 & 743 & 1,702 & 816 & 1,623 & 1,144 & 1,187 \\
Phi-4-Reasoning & 1,259 & 1,480 & 1,574 & 2,050 & 1,641 & 2,582 & 1,692 & 2,607 & 2,104 & 1,966 \\
\bottomrule
\end{tabular}}
\caption{Average token counts for base models across all datasets and models.}
\label{tab:base_token_counts_split}
\end{table*}

\begin{table*}[t]
\centering
\small
\setlength{\tabcolsep}{3pt}
\renewcommand{\arraystretch}{1.05}
\begin{tabular}{lcccccccccccccc}
\toprule
& \multicolumn{2}{c}{GSM8K} & \multicolumn{2}{c}{MATH500} & \multicolumn{2}{c}{AIME2024} & \multicolumn{2}{c}{AIME2025} & \multicolumn{2}{c}{GPQA-D} & \multicolumn{2}{c}{ARC-C} & \multicolumn{2}{c}{Avg.} \\
\cmidrule(lr){2-3}\cmidrule(lr){4-5}\cmidrule(lr){6-7}\cmidrule(lr){8-9}\cmidrule(lr){10-11}\cmidrule(lr){12-13}\cmidrule(lr){14-15}
Model
& Acc & $\Delta$Tok & Acc & $\Delta$Tok & Acc & $\Delta$Tok & Acc & $\Delta$Tok & Acc & $\Delta$Tok & Acc & $\Delta$Tok & Acc & $\Delta$Tok \\
\midrule
Qwen3-4B-Thinking & 95.1 & -- & 95.3 & -- & 77.1 & -- & 76.2 & -- & 64.1 & -- & 94.3 & -- & 83.7 & -- \\
\quad + NoWait & 94.2 & -25.1\% & 96.4 & -17.9\% & 66.7 & -30.5\% & 66.7 & -35.6\% & 68.2 & -19.6\% & 92.7 & 14.6\% & 80.8 & -19.0\% \\
\quad + ThinkLess & 92.6 & -100\% & 56.4 & -100\% & 3.3 & -100\% & 3.3 & -100\% & 50.0 & -100\% & 94.2 & -100\% & 50.0 & -100\% \\

\quad + DEER & 95.1 & -34.1\% & 94.4 & -14.2\% & 63.3 & -34.3\% & 60.0 & -37.9\% & 68.2 & -1.3\% & 92.8 & -9.0\% & 79.0 & -21.8\% \\
\quad + Dynasor-CoT & 95.2 & -34.0\% & 87.2 & -42.1\% & 63.3 & -28.8\% & 53.3 & -28.9\% & 62.6 & -22.6\% & 94.5 & -39.1\% & 76.0 & -23.0\% \\
\quad + \method{}-p & 93.9 & -37.3\% & 91.2 & -57.9\% & 35.0 & -72.6\% & 30.4 & -76.0\% & 47.0 & -67.5\% & 93.7 & -31.7\% & 65.2 & -57.2\% \\
\rowcolor{lightblue}
\quad + \method{} & 94.8 & -18.9\% & 96.6 & -20.4\% & 67.2 & -33.5\% & 62.8 & -33.4\% & 62.6 & -32.7\% & 94.0 & -19.9\% & 79.7 & -26.5\% \\
\midrule
Qwen3-4B & 94.4 & -- & 96.0 & -- & 68.8 & -- & 59.6 & -- & 51.0 & -- & 94.0 & -- & 77.3 & -- \\
\quad + NoWait & 94.5 & -40.4\% & 94.6 & -31.4\% & 56.7 & -16.6\% & 50.0 & -27.1\% & 57.6 & -26.6\% & 93.4 & -14.4\% & 74.5 & -26.1\% \\
\quad + ThinkLess & 91.0 & -100\% & 85.4 & -100\% & 23.3 & -100\% & 23.3 & -100\% & 46.0 & -100\% & 89.3 & -100\% & 59.7 & -100\% \\
\quad + \method{}-p & 92.8 & -40.2\% & 88.6 & -51.3\% & 29.2 & -64.5\% & 25.0 & -67.8\% & 47.5 & -54.2\% & 93.4 & -15.0\% & 62.8 & -48.8\% \\
\rowcolor{lightblue}
\quad + \method{} & 94.5 & -30.1\% & 95.4 & -14.4\% & 64.6 & -12.7\% & 60.4 & -13.5\% & 55.6 & -29.7\% & 93.9 & -11.5\% & 77.4 & -18.7\% \\
\midrule
Qwen3-14B & 96.0 & -- & 96.8 & -- & 71.7 & -- & 69.2 & -- & 60.6 & -- & 95.9 & -- & 81.7 & -- \\
\quad + NoWait & 95.8 & -29.4\% & 95.2 & -25.1\% & 70.0 & -19.8\% & 56.7 & -16.6\% & 62.1 & -20.5\% & 96.2 & -8.1\% & 79.3 & -19.9\% \\
\quad + ThinkLess & 93.4 & -100\% & 86.8 & -100\% & 23.3 & -100\% & 30.0 & -100\% & 49.5 & -100\% & 94.3 & -100\% & 62.9 & -100\% \\
\quad + \method{}-p & 95.1 & -31.6\% & 89.0 & -49.8\% & 38.8 & -68\% & 26.2 & -72\% & 57.6 & -66.4\% & 94.7 & -17.9\% & 66.9 & -51\% \\
\rowcolor{lightblue}
\quad + \method{} & 95.0 & -15.4\% & 96.9 & -7.7\% & 77.9 & -7.4\% & 65.8 & -6.0\% & 61.1 & -39.9\% & 95.2 & -18.2\% & 82.0 & -15.8\% \\
\midrule
Qwen3-32B & 96.0 & -- & 97.0 & -- & 75.8 & -- & 68.3 & -- & 65.2 & -- & 92.1 & -- & 82.4 & -- \\
\quad + NoWait & 95.3 & -23.6\% & 95.8 & -20.6\% & 66.7 & -12.0\% & 56.7 & -10.6\% & 60.1 & -12.4\% & 90.1 & 9.0\% & 77.5 & -11.7\% \\
\quad + ThinkLess & 93.3 & -100\% & 87.6 & -100\% & 23.3 & -100\% & 20.0 & -100\% & 54.5 & -100\% & 95.1 & -100\% & 62.3 & -100\% \\
\quad + \method{}-p & 94.8 & -23.0\% & 89.6 & -39.4\% & 39.2 & -61.5\% & 26.7 & -66.4\% & 53.5 & -51.0\% & 92.2 & -11.7\% & 66.0 & -42.2\% \\
\rowcolor{lightblue}
\quad + \method{} & 96.5 & -9.1\% & 97.2 & -1.4\% & 77.1 & -6.2\% & 68.3 & -4.3\% & 65.7 & -19.9\% & 91.0 & -8.0\% & 82.6 & -8.2\% \\
\midrule
DeepSeek-R1-7B & 92.7 & -- & 93.8 & -- & 49.5 & -- & 37.6 & -- & 48.0 & -- & 67.7 & -- & 64.9 & -- \\
\quad + NoWait & 90.6 & -30.8\% & 91.2 & -31.0\% & 40.0 & -32.4\% & 26.7 & -36.5\% & 43.4 & -35.5\% & 64.6 & -12.6\% & 59.4 & -29.8\% \\
\quad + ThinkLess & 85.9 & -100\% & 78.4 & -100\% & 16.7 & -100\% & 20.0 & -100\% & 30.3 & -100\% & 50.4 & -100\% & 47.0 & -100\% \\
\quad + DEER & 89.0 & -60.5\% & 87.2 & -44.5\% & 50.0 & -28.2\% & 33.3 & -9.8\% & 43.9 & -16.7\% & 73.9 & -10.6\% & 62.9 & -28.3\% \\
\quad + Dynasor-CoT & 61.3 & -87.8\% & 64.2 & -47.2\% & 33.4 & -31.8\% & 33.4 & -26.4\% & 32.3 & -71.2\% & 77.8 & -53.6\% & 50.4 & -53.0\% \\
\quad + \method{}-p & 87.8 & -54.9\% & 81.4 & -60.6\% & 25.4 & -71.4\% & 22.9 & -73.5\% & 36.4 & -76.8\% & 67.2 & -31.0\% & 53.5 & -61.4\% \\
\rowcolor{lightblue}
\quad + \method{} & 92.4 & -25.0\% & 92.0 & -23.2\% & 45.2 & -25.0\% & 33.2 & -20.3\% & 45.5 & -51.5\% & 70.7 & -35.3\% & 63.2 & -30.0\% \\
\midrule
Phi-4-Reasoning & 91.7 & -- & 71.1 & -- & 67.5 & -- & 63.7 & -- & 62.6 & -- & 80.0 & -- & 72.8 & -- \\
\quad + NoWait & 92.1 & -2.4\% & 70.8 & -17.7\% & 63.3 & -9.4\% & 53.3 & -8.6\% & 63.1 & -19.1\% & 77.6 & -7.0\% & 70.0 & -10.7\% \\
\quad + ThinkLess & 92.6 & -100\% & 66.4 & -100\% & 36.7 & -100\% & 23.3 & -100\% & 54.0 & -100\% & 82.7 & -100\% & 59.3 & -100\% \\
\quad + \method{}-p & 91.3 & -42.5\% & 68.8 & -50.4\% & 39.6 & -63.1\% & 29.2 & -66.6\% & 59.1 & -62.8\% & 84.2 & -49.2\% & 62.0 & -55.8\% \\
\rowcolor{lightblue}
\quad + \method{} & 91.7 & -25.6\% & 70.4 & -17.2\% & 73.3 & -6.4\% & 53.3 & -5.2\% & 67.7 & -1.1\% & 80.0 & -25.3\% & 72.7 & -13.5\% \\
\midrule
Qwen3-Next-80B & 96.7 & -- & 98.2 & -- & 90.0 & -- & 80.0 & -- & 76.3 & -- & 96.8 & -- & 89.7 & -- \\
\rowcolor{lightblue}
\quad + \method{} & 96.5 & -24.3\% & 99.2 & -21.6\% & 86.7 & -24.3\% & 73.3 & -16.7\% & 72.7 & -25.6\% & 97.0 & -21.8\% & 87.6 & -22.3\% \\
\bottomrule
\end{tabular}
\caption{Math and science results on GSM8K, MATH500, AIME24, AIME25, GPQA-D, and ARC-C. We report accuracy and $\Delta$Tok (thinking token reduction vs. the Base decoding) for each model. This table presents the full comparison including all baseline methods for all models.}
\label{tab:math_results_full}
\end{table*}

\clearpage

\begin{table*}[t]
\centering
\small
\setlength{\tabcolsep}{1.5pt}
\renewcommand{\arraystretch}{1.05}
\makebox[\textwidth][c]{ 
\resizebox{\textwidth}{!}{
\begin{tabular}{lcccccccccccccccccccccc}
\toprule
& \multicolumn{8}{c}{BFCL-v1} & \multicolumn{8}{c}{BFCL-v2} & \multicolumn{4}{c}{Meta-Tool} & \multicolumn{2}{c}{} \\ \cmidrule(lr){2-9}\cmidrule(lr){10-17}\cmidrule(lr){18-21} Model & \multicolumn{2}{c}{Simple} & \multicolumn{2}{c}{Multi} & \multicolumn{2}{c}{Parallel} & \multicolumn{2}{c}{Multi-Par.} & \multicolumn{2}{c}{Simple} & \multicolumn{2}{c}{Multi} & \multicolumn{2}{c}{Parallel} & \multicolumn{2}{c}{Multi-Par.} & \multicolumn{2}{c}{Single} & \multicolumn{2}{c}{Multiple} & \multicolumn{2}{c}{Avg.} \\ \cmidrule(lr){2-3}\cmidrule(lr){4-5}\cmidrule(lr){6-7}\cmidrule(lr){8-9}\cmidrule(lr){10-11}\cmidrule(lr){12-13}\cmidrule(lr){14-15}\cmidrule(lr){16-17}\cmidrule(lr){18-19}\cmidrule(lr){20-21}\cmidrule(lr){22-23} & Acc & $\Delta$Tok & Acc & $\Delta$Tok & Acc & $\Delta$Tok & Acc & $\Delta$Tok & Acc & $\Delta$Tok & Acc & $\Delta$Tok & Acc & $\Delta$Tok & Acc & $\Delta$Tok & Acc & $\Delta$Tok & Acc & $\Delta$Tok & Acc & $\Delta$Tok \\
\midrule
Qwen3-4B-Thinking & 88.2 & -- & 90.5 & -- & 93.5 & -- & 90.5 & -- & 79.8 & -- & 83.4 & -- & 87.5 & -- & 79.2 & -- & 69.7 & -- & 85.5 & -- & 84.8 & -- \\
\quad + NoWait & 83.6 & -33.1\% & 88.0 & -11.5\% & 88.5 & -44.4\% & 88.5 & -53.5\% & 75.6 & -29.9\% & 80.3 & -16.5\% & 81.3 & -50.7\% & 75.0 & -66.5\% & 68.9 & 0.1\% & 84.7 & -11.2\% & 81.4 & -31.7\% \\
\quad + ThinkLess & 86.2 & -8.1\% & 93.0 & -2.2\% & 79.5 & -100.0\% & 83.0 & -100.0\% & 81.0 & -5.9\% & 82.9 & -4.1\% & 37.5 & -100.0\% & 45.8 & -100.0\% & 67.4 & -100.0\% & 86.5 & -100.0\% & 74.3 & -62.0\% \\
\quad + \method{}-p & 88.0 & -40.0\% & 90.5 & -16.7\% & 78.0 & -26.9\% & 42.5 & -33.4\% & 79.5 & -34.8\% & 81.7 & -20.4\% & 62.5 & -28.5\% & 29.2 & -43.8\% & 70.8 & -25.8\% & 86.7 & -12.7\% & 70.9 & -28.3\% \\
\rowcolor{lightblue}
\quad + \method{} & 87.3 & -32.9\% & 90.0 & -12.4\% & 95.5 & -26.8\% & 90.5 & -28.2\% & 82.2 & -29.9\% & 82.9 & -16.7\% & 87.5 & -32.3\% & 87.5 & -18.9\% & 72.4 & -20.6\% & 86.3 & -9.0\% & 86.2 & -22.8\% \\
\midrule
Qwen3-4B & 88.2 & -- & 92.0 & -- & 90.0 & -- & 83.0 & -- & 80.2 & -- & 81.0 & -- & 81.2 & -- & 70.8 & -- & 74.3 & -- & 90.5 & -- & 83.1 & -- \\
\quad + NoWait & 88.2 & -16.3\% & 90.5 & -1.5\% & 88.0 & -13.5\% & 84.0 & 1.2\% & 78.3 & -9.6\% & 80.0 & -5.8\% & 75.0 & -5.1\% & 75.0 & -11.1\% & 72.5 & -3.1\% & 90.5 & -8.4\% & 82.2 & -7.3\% \\
\quad + ThinkLess & 86.4 & -100.0\% & 89.5 & -100.0\% & 87.5 & -100.0\% & 83.5 & -100.0\% & 71.3 & -100.0\% & 76.4 & -100.0\% & 56.2 & -100.0\% & 62.5 & -100.0\% & 69.1 & -100.0\% & 92.2 & -100.0\% & 77.5 & -100.0\% \\
\quad + \method{}-p & 87.3 & -13.5\% & 93.0 & -4.6\% & 90.5 & -14.8\% & 84.0 & -4.8\% & 80.2 & -6.9\% & 81.8 & -3.5\% & 75.0 & -15.1\% & 75.0 & -12.7\% & 73.5 & -1.5\% & 90.3 & -2.9\% & 83.1 & -8.0\% \\
\rowcolor{lightblue}
\quad + \method{} & 87.8 & -12.5\% & 92.5 & -5.2\% & 89.5 & -16.5\% & 83.5 & -12.0\% & 79.1 & -9.0\% & 81.2 & -2.9\% & 62.5 & -21.1\% & 75.0 & -10.2\% & 72.7 & -1.4\% & 91.1 & -3.0\% & 81.5 & -9.4\% \\
\midrule
Qwen3-14B & 87.8 & -- & 93.0 & -- & 92.0 & -- & 83.5 & -- & 80.6 & -- & 81.1 & -- & 62.5 & -- & 70.8 & -- & 63.3 & -- & 84.9 & -- & 80.0 & -- \\
\quad + NoWait & 87.1 & -7.5\% & 92.0 & -0.6\% & 90.0 & -16.3\% & 84.0 & 1.8\% & 77.9 & -7.0\% & 79.8 & -4.9\% & 62.5 & -30.1\% & 62.5 & -10.9\% & 61.2 & -2.4\% & 84.5 & -5.8\% & 78.2 & -8.4\% \\
\quad + ThinkLess & 88.4 & -100.0\% & 87.5 & -100.0\% & 90.0 & -100.0\% & 87.0 & -100.0\% & 74.4 & -100.0\% & 77.9 & -100.0\% & 56.2 & -100.0\% & 50.0 & -100.0\% & 69.7 & -100.0\% & 83.5 & -100.0\% & 76.5 & -100.0\% \\
\quad + \method{}-p & 88.4 & -9.4\% & 92.5 & -2.2\% & 91.5 & -23.5\% & 86.0 & -16.0\% & 77.9 & -10.1\% & 80.7 & -5.3\% & 68.8 & -29.9\% & 50.0 & -14.6\% & 64.6 & -2.1\% & 85.9 & -4.8\% & 78.6 & -11.8\% \\
\rowcolor{lightblue}
\quad + \method{} & 87.8 & -9.5\% & 92.0 & -4.1\% & 93.0 & -21.2\% & 85.0 & -14.2\% & 78.3 & -7.7\% & 79.7 & -5.2\% & 62.5 & -25.4\% & 62.5 & -7.6\% & 66.3 & -1.6\% & 85.7 & -4.5\% & 79.3 & -10.1\% \\
\midrule
Qwen3-32B & 88.0 & -- & 90.0 & -- & 93.0 & -- & 85.0 & -- & 83.7 & -- & 80.7 & -- & 68.8 & -- & 70.8 & -- & 64.8 & -- & 84.3 & -- & 80.9 & -- \\
\quad + NoWait & 87.3 & -7.8\% & 88.0 & -4.0\% & 91.0 & -13.0\% & 82.5 & -5.0\% & 80.6 & -6.2\% & 79.0 & -1.6\% & 68.8 & -3.9\% & 62.5 & -14.3\% & 63.8 & 3.6\% & 83.9 & 1.3\% & 78.7 & -5.1\% \\
\quad + ThinkLess & 89.1 & -100.0\% & 87.5 & -100.0\% & 91.5 & -100.0\% & 88.0 & -100.0\% & 77.1 & -100.0\% & 81.3 & -100.0\% & 68.8 & -100.0\% & 58.3 & -100.0\% & 72.3 & -100.0\% & 82.7 & -100.0\% & 79.7 & -100.0\% \\
\quad + \method{}-p & 89.1 & -0.5\% & 91.5 & -3.2\% & 92.0 & -12.3\% & 84.5 & -7.1\% & 80.2 & -6.2\% & 80.8 & -0.7\% & 75.0 & -9.1\% & 66.7 & -15.6\% & 64.4 & -1.6\% & 85.1 & -3.3\% & 80.9 & -6.0\% \\
\rowcolor{lightblue}
\quad + \method{} & 88.9 & -5.4\% & 91.0 & -3.7\% & 91.0 & -14.3\% & 90.0 & -7.3\% & 79.8 & -7.8\% & 81.5 & -2.2\% & 68.8 & -4.6\% & 58.0 & -13.3\% & 64.4 & -1.5\% & 85.5 & -1.1\% & 79.9 & -6.1\% \\
\midrule
DeepSeek-R1-7B & 57.8 & -- & 61.0 & -- & 52.5 & -- & 38.0 & -- & 54.7 & -- & 41.7 & -- & 43.8 & -- & 29.2 & -- & 63.2 & -- & 80.3 & -- & 52.2 & -- \\
\quad + NoWait & 63.3 & -2.2\% & 65.5 & 0.7\% & 62.0 & -4.0\% & 42.5 & -6.9\% & 57.8 & -2.0\% & 39.6 & -3.2\% & 50.0 & -8.9\% & 20.8 & 11.0\% & 64.9 & 0.6\% & 73.2 & -19.3\% & 54.0 & -3.4\% \\
\quad + ThinkLess & 9.6 & -100.0\% & 23.5 & -100.0\% & 5.0 & -100.0\% & 5.0 & -100.0\% & 24.0 & -100.0\% & 20.1 & -100.0\% & 0.0 & -100.0\% & 4.2 & -100.0\% & 57.4 & -100.0\% & 88.3 & -100.0\% & 23.7 & -100.0\% \\
\quad + \method{}-p & 48.4 & -9.7\% & 50.0 & -9.1\% & 39.0 & -12.1\% & 28.5 & -11.6\% & 46.5 & -8.3\% & 35.8 & -7.7\% & 37.5 & -5.8\% & 20.8 & -5.2\% & 65.0 & -9.3\% & 79.5 & -26.3\% & 45.1 & -10.5\% \\
\rowcolor{lightblue}
\quad + \method{} & 45.3 & -11.4\% & 50.5 & -9.6\% & 42.0 & -14.6\% & 37.0 & -11.6\% & 46.1 & -10.8\% & 35.8 & -8.8\% & 43.8 & -19.9\% & 16.7 & -6.0\% & 63.4 & -7.4\% & 76.3 & -26.3\% & 45.7 & -12.6\% \\
\midrule
Phi-4-Reasoning & 82.7 & -- & 88.0 & -- & 87.5 & -- & 77.5 & -- & 70.9 & -- & 72.6 & -- & 81.2 & -- & 58.3 & -- & 77.7 & -- & 90.5 & -- & 78.7 & -- \\
\quad + NoWait & 79.1 & -6.8\% & 83.0 & -11.1\% & 81.0 & -1.3\% & 75.5 & -13.2\% & 70.5 & -16.7\% & 67.5 & -12.5\% & 81.2 & -6.4\% & 58.3 & -11.9\% & 76.6 & -8.5\% & 90.9 & -15.4\% & 76.4 & -10.4\% \\
\quad + ThinkLess & 65.8 & -100.0\% & 72.5 & -100.0\% & 73.5 & -100.0\% & 68.5 & -100.0\% & 56.2 & -100.0\% & 66.3 & -100.0\% & 87.5 & -100.0\% & 50.0 & -100.0\% & 76.3 & -100.0\% & 90.5 & -100.0\% & 70.7 & -100.0\% \\
\quad + \method{}-p & 80.4 & -30.0\% & 85.5 & -22.0\% & 88.5 & -37.8\% & 80.0 & -39.1\% & 62.8 & -36.2\% & 62.8 & -26.4\% & 87.5 & -42.8\% & 50.0 & -28.7\% & 75.2 & -36.1\% & 91.3 & -43.8\% & 76.4 & -34.3\% \\
\rowcolor{lightblue}
\quad + \method{} & 81.6 & 0.9\% & 87.5 & -4.2\% & 84.5 & 2.7\% & 77.5 & -6.4\% & 69.4 & -8.1\% & 72.2 & -1.9\% & 75.0 & 4.6\% & 70.8 & -4.9\% & 77.2 & -2.5\% & 90.3 & -5.4\% & 78.6 & -2.5\% \\
\bottomrule
\end{tabular}}}
\caption{Results on BFCL-v1, BFCL-v2, and Meta-Tool benchmarks. We report accuracy and $\Delta$Tok for all subcategories (Simple, Multiple, Parallel, Multi-Parallel for BFCL; Single, Multiple for Meta-Tool) across all methods.}
\label{tab:bfcl_meta_results_full}
\end{table*}

\begin{table}[t]
\centering
\small
\begin{tabular}{lllcccc}
\toprule
Model & Metric & Statistics
& \multicolumn{2}{c}{AIME2024}
& \multicolumn{2}{c}{AIME2025} \\
\cmidrule(lr){4-5} \cmidrule(lr){6-7}
 &  & 
& Baseline & \method{}
& Baseline & \method{} \\
\midrule

\multirow{4}{*}{Qwen3-4B-Thinking}
& \multirow{2}{*}{Accuracy}
& Variance 
& 18.2 & 20.2 & 23.7 & 43.1 \\
&  & 95\% CI 
& {\footnotesize [76.2, 79.2]} & {\footnotesize [65.6, 68.7]}
& {\footnotesize [74.5, 77.8]} & {\footnotesize [60.5, 65.1]} \\

\cmidrule(lr){2-7}

& \multirow{2}{*}{Tokens}
& Variance 
& 78,749,681 & 33,023,520 & 74,600,230 & 37,478,663 \\
&  & 95\% CI 
& {\footnotesize [18,870, 19,994]} & {\footnotesize [12,605, 13,333]}
& {\footnotesize [20,725, 21,819]} & {\footnotesize [13,727, 14,502]} \\

\midrule

\multirow{4}{*}{DeepSeek-R1-7B}
& \multirow{2}{*}{Accuracy}
& Variance 
& 44.5 & 33.5 & 22.4 & 30.5 \\
&  & 95\% CI 
& {\footnotesize [47.2, 51.8]} & {\footnotesize [43.4, 47.4]}
& {\footnotesize [36.0, 39.2]} & {\footnotesize [31.3, 35.1]} \\

\cmidrule(lr){2-7}

& \multirow{2}{*}{Tokens}
& Variance 
& 27,202,081 & 17,728,953 & 28,471,392 & 19,546,656 \\
\addlinespace[0.3em]
&  & 95\% CI 
& {\footnotesize [10,420, 11,081]} & {\footnotesize [7,885, 8,418]}
& {\footnotesize [10,872, 11,548]} & {\footnotesize [8,173, 8,733]} \\

\bottomrule
\end{tabular}
\caption{Variance and 95\% confidence intervals (CI) of accuracy and token count.}
\label{tab:variance_ci_results}
\end{table}

\begin{table*}[t]
\centering
\small
\setlength{\tabcolsep}{3pt}
\renewcommand{\arraystretch}{1.05}
\resizebox{\textwidth}{!}{
\begin{tabular}{lccccc}
\toprule
Model & Qwen3-4B-Thinking & Qwen3-4B & Qwen3-14B & Qwen3-32B & Qwen3-Next-80B \\
\midrule
Avg. Base Token &	9,695	& 7,352	& 6,976	& 6,634	& 7,844 \\
$\Delta\text{Tok}$ &	-22.6\% &	-10.7\% & 	-12.4\% &	-7.0\% &	-22.3\% \\
\bottomrule
\end{tabular}}
\caption{Token reduction rate of \method{}. We report $\Delta$Tok (thinking token reduction vs. the Base decoding) for Qwen3 Family. }
\label{tab:reasoning_trace_trend_results}
\end{table*}

\clearpage

\begin{table*}[t]
\centering
\begin{tabular}{ccrrr}
\toprule
Benchmark & Problem ID & Oracle & \method{} & $\Delta\text{Tok}$ \\
\midrule
\multirow{7}{*}{AIME2024}
 & 1  & 5,734  & 21,555 & {\color{green!60!black}+15,821} \\
 & 13 & 4,962  & 11,244 & {\color{green!60!black}+6,282} \\
 & 18 & 3,948  & 21,967 & {\color{green!60!black}+18,019} \\
 & 20 & 15,969 & 21,394 & {\color{green!60!black}+5,425} \\
 & 25 & 6,778  & 22,777 & {\color{green!60!black}+15,999} \\
 & 27 & 4,396  & 14,965 & {\color{green!60!black}+10,569} \\
 & 29 & 17,396 & 10,685 & {\color{red}-6,711} \\
\midrule
\multirow{9}{*}{AIME2025}
 & 9  & 20,393 & 10,190 & {\color{red}-10,203} \\
 & 10 & 7,660  & 13,672 & {\color{green!60!black}+6,012} \\
 & 12 & 6,369  & 21,539 & {\color{green!60!black}+15,170} \\
 & 13 & 6,528  & 21,690 & {\color{green!60!black}+15,162} \\
 & 19 & 2,730  & 12,770 & {\color{green!60!black}+10,040} \\
 & 25 & 8,097  & 21,605 & {\color{green!60!black}+13,508} \\
 & 27 & 11,175 & 22,880 & {\color{green!60!black}+11,705} \\
 & 28 & 6,490  & 22,424 & {\color{green!60!black}+15,934} \\
 & 29 & 4,391  & 21,641 & {\color{green!60!black}+17,250} \\
\bottomrule
\end{tabular}
\caption{Token comparison (Oracle vs. \method{}) on AIME2024 and AIME2025 for Qwen3-4B-Thinking.}
\label{tab:oracle_thinkbrake_token_delta}
\end{table*}

\vspace*{\fill}

\begin{longtable}{p{0.96\textwidth}}
\toprule
\small\ttfamily
\endfirsthead
\toprule
\small\ttfamily
\endhead
\bottomrule
\endfoot
\endlastfoot
\texttt{<think>} \\
... \\

\hlg{But wait, in the problem, they say "the circumcircle of triangle DEF intersects BD, AE, AF at G, H, J". Wait, BD: B to D, D is midpoint of BC, so BD is a median? Wait, no, median from B should go to midpoint of AC, but here D is midpoint of BC, so BD is not a median, it's a segment from vertex B to midpoint of side BC? Wait, that's just half of side BC! Wait, hold on, that can't be. Wait, maybe I misread the problem. Wait, let's check again: "D, E, and F be the midpoints of sides BC, AC, and AB, respectively." So yes: D $\in$ BC, midpoint; E $\in$ AC, midpoint; F $\in$ AB, midpoint.}\\
\texttt{// Oracle Stopping point} \\
\\

\hlr{Then "the circumcircle of triangle DEF intersects BD, AE, and AF at points G, H, J, respectively." Wait, BD: B is a vertex, D is midpoint of BC, so BD is the segment from B to D, which is half of BC?}\\
\texttt{// Spurious Reasoning begins} \\

... \\

\hlr{Yes! This makes perfect sense now. The nine-point circle passes through the three feet of the altitudes and the three midpoints of the sides, so each side of the triangle intersects the nine-point circle at exactly two points: the midpoint of the side and the foot of the altitude from the opposite vertex.}\\
\texttt{// False Confidence, \method{} stopping leads to wrong answer}\\

... \\

\hlg{I can't believe I didn't remember that the nine-point circle passes through the feet of the altitudes earlier---}\\
\texttt{// Correction after continued Reasoning}\\

\bottomrule
\caption{Spurious reasoning example from AIME2025-19. The model initially reaches a correct interpretation (oracle stopping point), but continues spurious reasoning, leading \method{} to terminate at a false-confidence state.}

\label{tab:spurious_trace}
\end{longtable}

\clearpage

\twocolumn

\end{document}